\documentclass[10pt,twocolumn,letterpaper]{article}

\usepackage[pagenumbers]{cvpr} 
\usepackage{amssymb}
\usepackage{pifont}
\usepackage{tabularray}
\definecolor{deepgreen}{rgb}{0.0, 0.5, 0.0}  

\definecolor{cvprblue}{rgb}{0.21,0.49,0.74}
\usepackage[pagebackref,breaklinks,colorlinks,allcolors=cvprblue]{hyperref}
\usepackage{enumitem}
\usepackage{pifont}
\usepackage{subcaption}
\usepackage{arydshln}

\newcommand{\ourmodel}{AerOSeg}

\def\paperID{24} 
\def\confName{EarthVision}
\def\confYear{2025}

\title{AerOSeg: Harnessing SAM for Open-Vocabulary Segmentation in Remote Sensing Images}

\author{Saikat Dutta\textsuperscript{1,2,3}\thanks{C-MInDS, IIT Bombay --- \texttt{23d2031@iitb.ac.in}}
\qquad
Akhil Vasim\textsuperscript{2}
\qquad
Siddhant Gole\textsuperscript{2}
\\
Hamid Rezatofighi\textsuperscript{3}
\qquad 
Biplab Banerjee\textsuperscript{2}
\\[5pt]
\textsuperscript{1}IITB-Monash Research Academy \quad \textsuperscript{2}IIT Bombay \quad \textsuperscript{3}Monash University
}

\begin{document}
\maketitle
\begin{abstract}

Image segmentation beyond predefined categories is a key challenge in remote sensing, where novel and unseen classes often emerge during inference. Open-vocabulary image Segmentation addresses these generalization issues in traditional supervised segmentation models while reducing reliance on extensive per-pixel annotations, which are both expensive and labor-intensive to obtain. Most Open-Vocabulary Segmentation (OVS) methods are designed for natural images but struggle with remote sensing data due to scale variations, orientation changes, and complex scene compositions. This necessitates the development of OVS approaches specifically tailored for remote sensing. In this context, we propose \textbf{AerOSeg}, a novel OVS approach for remote sensing data. First, we compute robust image-text correlation features using multiple rotated versions of the input image and domain-specific prompts. These features are then refined through spatial and class refinement blocks. Inspired by the success of the Segment Anything Model (SAM) in diverse domains, we leverage SAM features to guide the spatial refinement of correlation features. Additionally, we introduce a semantic back-projection module and loss to ensure the seamless propagation of SAM’s semantic information throughout the segmentation pipeline. Finally, we enhance the refined correlation features using a multi-scale attention-aware decoder to produce the final segmentation map. We validate our SAM-guided Open-Vocabulary Remote Sensing Segmentation model on three benchmark remote sensing datasets: iSAID, DLRSD, and OpenEarthMap. Our model outperforms state-of-the-art open-vocabulary segmentation methods, achieving an average improvement of 2.54 h-mIoU.
\end{abstract}

\vspace{-20px}
\section{Introduction}
\label{sec:intro}

\begin{figure}[!t]
    \centering
    \includegraphics[width=\linewidth]{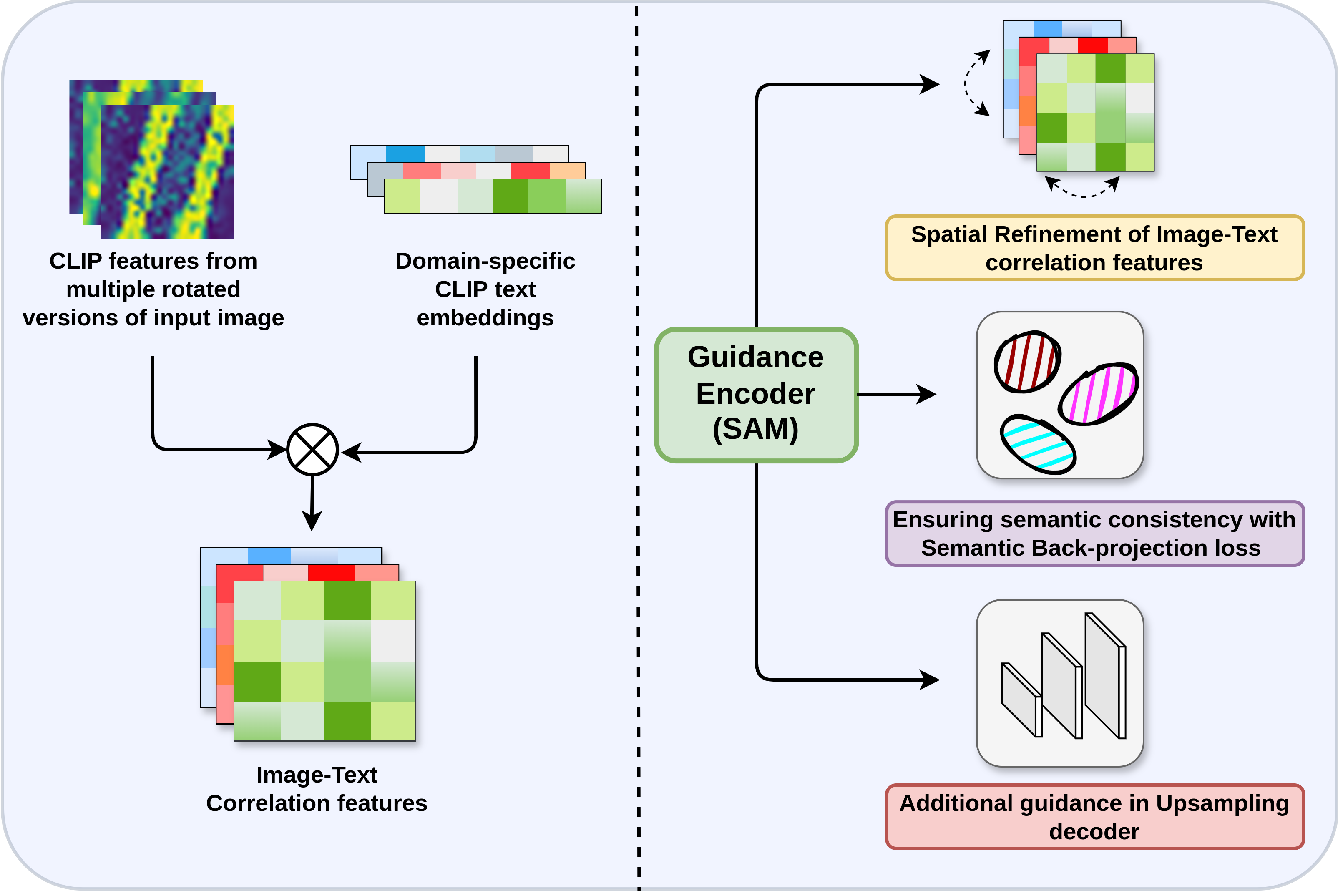}
    \caption{Key aspects of \textbf{\ourmodel}: (a) Correlation features computation from CLIP features across rotated inputs and domain-specific text embeddings. (b) Guidance Encoder's multifaceted contribution.}
    \vspace{-15px}
    \label{fig:teaser}
\end{figure}

Open-vocabulary image segmentation aims to segment objects belonging to an unbounded set of categories, thereby allowing the model to handle novel or previously unseen classes at inference. Unlike traditional semantic segmentation --- where the training and testing categories are strictly identical --- open-vocabulary segmentation endeavors to learn transferable semantic representations that can facilitate the segmentation of any class not observed during training. In the context of \emph{remote sensing} imagery, this capability becomes especially critical: supervised segmentation approaches typically require large-scale datasets with pixel-level annotations that are notoriously expensive and time-consuming to obtain, primarily due to high spatial resolution, complex scenes, and varying geospatial contexts. Moreover, models trained solely on these well-annotated classes tend to overfit, limiting their ability to generalize to new or rare categories that often arise in real-world remote sensing applications (e.g., emerging structures, seasonal objects, or natural disasters).

Most recent open-vocabulary segmentation methods leverage vision-language models such as CLIP \cite{clip} or ALIGN \cite{align}, which learn powerful joint embeddings of visual and textual data. Despite their success with natural images, these models often struggle with satellite imagery, where objects can appear at drastically different scales, orientations, and resolutions. Remote sensing images are fundamentally distinct from typical natural images due to large scene coverage, small object sizes, and complex backgrounds, necessitating specialized strategies that address these domain-specific challenges.

Motivated by these requirements, researchers have recently proposed open-vocabulary solutions tailored to remote sensing. A pioneering approach, Open-Vocabulary Remote Sensing Segmentation (OVRS) \cite{cao2024ovrs}, introduces a rotational invariance paradigm by augmenting the visual encoder input with multiple rotated versions of the image, thereby capturing orientation-invariant semantic features. Subsequent modules refine these features spatially and by category and finally leverage a multi-scale decoder for high-resolution segmentation. While effective, OVRS heavily relies on CLIP for open-vocabulary segmentation, which is suboptimal. CLIP, while excelling in open-vocabulary classification, is designed to align global visual and textual semantics, limiting its effectiveness for per-pixel tasks like segmentation.

In this work, we hypothesize that \textbf{complementary guidance features} can help preserve semantic richness throughout the refinement pipeline. Specifically, we propose exploiting the \textbf{Segment Anything Model (SAM)} as a feature guidance encoder, in conjunction with CLIP, to strengthen the semantic flow in open-vocabulary remote sensing segmentation. 
SAM, trained on large-scale data for promptable segmentation, produces semantically rich features and performs well across diverse domains and datasets. Given the smaller size of remote sensing datasets, leveraging SAM's features provides valuable additional guidance for improved segmentation.
Our framework, \textbf{\ourmodel}, first computes image–text correlation maps from multiple rotated versions of the input image and specialized textual prompts, ensuring orientation-invariant features. Instead of generic prompts, we use domain-specific Remote Sensing prompts, enhancing the model’s robustness.
A swin transformer-based spatial refinement block, guided by SAM features, then refines these correlation maps, followed by a class refinement block that further sharpens category-specific responses. To address the loss of semantic fidelity after repeated refinements, we introduce a back-projection module that reproduces SAM features and enforces feature alignment via a \emph{semantic back-projection loss}, ensuring that semantic content is preserved. Finally, the refined correlation maps are upscaled via an Attention-aware Upsampling decoder to produce the final segmentation predictions. The salient aspects of \ourmodel\ is presented in Fig.~\ref{fig:teaser}.

In summary, our main contributions are as follows:
\begin{itemize}
    \item \textbf{SAM as Guidance:} We propose a novel synergy between SAM and CLIP for remote sensing segmentation, utilizing SAM as a feature guidance encoder to enrich semantic representations and mitigate loss of generalizability.
    \item \textbf{Back-Projection for Semantic Fidelity:} We integrate a back projection module and a corresponding feature reconstruction loss to sustain crucial semantic information derived from SAM, preventing over-refinement that impairs generalization to unseen classes.
    \item \textbf{Comprehensive Experiments:} We conduct extensive evaluations on three benchmark remote sensing segmentation datasets, demonstrating that our approach significantly outperforms state-of-the-art open-vocabulary segmentation methods on both seen and unseen categories.
\end{itemize}

\section{Related Works}

\noindent \textbf{A. Semantic segmentation in remote sensing:}
Image segmentation partitions an image into regions corresponding to specific classes or objects through pixel-wise grouping. In remote sensing, the high resolution and complexity of images present significant challenges. Modern deep learning methods, particularly Convolutional Neural Networks (CNNs), have advanced this field by learning intricate features from large-scale images \cite{rs13163054, ssunc, YUAN2021114417}. Building on CNNs, subsequent work has addressed class imbalance with weighted uncertainty labeling \cite{BRESSAN2022102690}, enabled multi-scale feature learning via pyramid attention pooling \cite{article}, and enhanced spatial detail capture through global context integration \cite{rs13010071}. Despite these advances, CNNs are inherently limited to local feature extraction.

Vision Transformers (ViTs) have emerged as a powerful alternative by leveraging attention mechanisms to model long-range dependencies \cite{transformers}. This has led to their increasing adoption in remote sensing segmentation. For example, Xu et al. \cite{rs13183585} introduced an Efficient Transformer backbone, adapted from the Swin Transformer \cite{liu2021swin}, which integrates an MLP head and auxiliary edge fusion to reduce computational costs and improve edge segmentation. However, due to the heavy computational burden of full attention mechanisms, most recent approaches combine ViTs and CNNs. Hybrid models, such as CVMH-Unet \cite{cao2024remotesensingimagesegmentation} and CCTNet \cite{rs14091956}, effectively merge global and local feature extraction while maintaining efficiency. Encoder-decoder hybrids like UNetFormer \cite{Wang_2022} and frameworks employing Swin Transformer encoders with CNN decoders \cite{9686732} further illustrate the trend toward integrating multi-scale feature representations in remote sensing segmentation. Despite achieving decent performance on benchmark datasets, this class of segmentation models can not be extended to segment an arbitrary number of classes during deployment.  

\noindent \textbf{B. Open-Vocabulary segmentation:}
Open-vocabulary segmentation is a challenging task that seeks to segment objects from an open set of categories defined by textual labels or descriptions. Unlike conventional segmentation, the label sets during training and testing can differ significantly.

Early methods focused on aligning visual embeddings with pre-defined word embeddings \cite{zs3net, xian2019semantic}. More recent approaches leverage large-scale visual-language models to align visual and semantic feature spaces. For instance, Li et al. \citep{lseg} integrate CLIP’s \cite{clip} text encoder with a DPT-based image encoder \cite{dpt}, employing a contrastive loss to align pixel embeddings with corresponding textual embeddings. Similarly, Ghiasi et al. \cite{openseg} introduce a class-agnostic segmentation module based on region-to-image cross-attention \cite{maxdeeplab} that generates segmentation masks used for visual-semantic alignment. In the same vein, Zegformer \cite{zegformer} decouples the task into Maskformer-based class-agnostic segmentation \cite{maskformer} and a zero-shot classification of segments using CLIP embeddings, while Xu et al. \cite{odise} exploit internal feature representations from text-to-image diffusion models to predict segmentation masks, classifying segments via CLIP text embeddings.

Additional contributions include the Side Adapter Network \cite{san}, which transforms images into visual tokens, appends query tokens, and integrates CLIP features within transformer layers. These are then processed by MLP layers to produce attention biases and mask proposals. Yu et al. \cite{fcclip} demonstrate that a frozen convolutional CLIP backbone can robustly perform open-vocabulary classification and mask generation, even at higher image resolutions. Cho et al. \cite{catseg} refine a cost-volume based on cosine similarity between CLIP image and text embeddings to generate segmentation maps. Shan et al. \cite{ebseg} perform elementwise addition of frozen CLIP and SAM features, which are then processed by a transformer-based decoder to generate segmentation maps through embedding balancing. Wang et al. \cite{wang2024use} introduce a data pipeline for curating extensive segment-text pairs alongside a universal segment embedding model to classify segments into diverse text-defined categories.

Although most research in open-vocabulary segmentation has targeted natural images, only a few works extend to other domains such as remote sensing. For example, Cao et al. \cite{cao2024ovrs} compute feature correlations between rotated image embeddings and text embeddings and utilize an attention-aware upsampling decoder to generate segmentation outputs. Similarly, Li et al. \cite{li2024segearth} employ a feature upsampler to enhance low-resolution features for training-free open-vocabulary segmentation in remote sensing imagery.

\noindent \textbf{C. Segment Anything Model:} The Segment Anything Model (SAM) \citep{sam} is a zero-shot segmentation framework trained on 1.1 billion masks and 11 million images. Given an image and a visual prompt—such as a bounding box, point annotations, or an initial mask—SAM employs an image encoder and a prompt encoder to generate corresponding embeddings. These embeddings are fused in a lightweight mask decoder to predict segmentation masks, ensuring robust outputs even when prompts are ambiguous.

SAM has demonstrated impressive interactive segmentation performance across diverse domains, including medical imaging, agriculture, food, and remote sensing. Recent work has sought to enhance and extend SAM's capabilities by developing domain-specific variants \cite{wu2023medicalsam, wang2024samrs}, incorporating additional prompt modalities \cite{zou2024seem}, improving computational efficiency, and even adapting the model for promptable video object segmentation \cite{ravi2024sam2}. Inspired by the generalization ability of SAM in multiple domains, we utilize features extracted by the SAM encoder to refine CLIP image-text correlation features in this work.

\vspace{-5px}

\section{Proposed Methodology}

\vspace{-5px}
\textbf{Problem Definition:} Given an input Remote Sensing Image $I$, the aim is to classify each pixel  from a set of categories defined by textual labels or description. Different from traditional set-up, class-set during training $\mathcal{C}_{train}$ can be different from  class-set during inference, $\mathcal{C}_{test}$ in Open-vocabulary setup. 

\begin{figure*}[!htp]
    \centering
    \includegraphics[width=\linewidth]{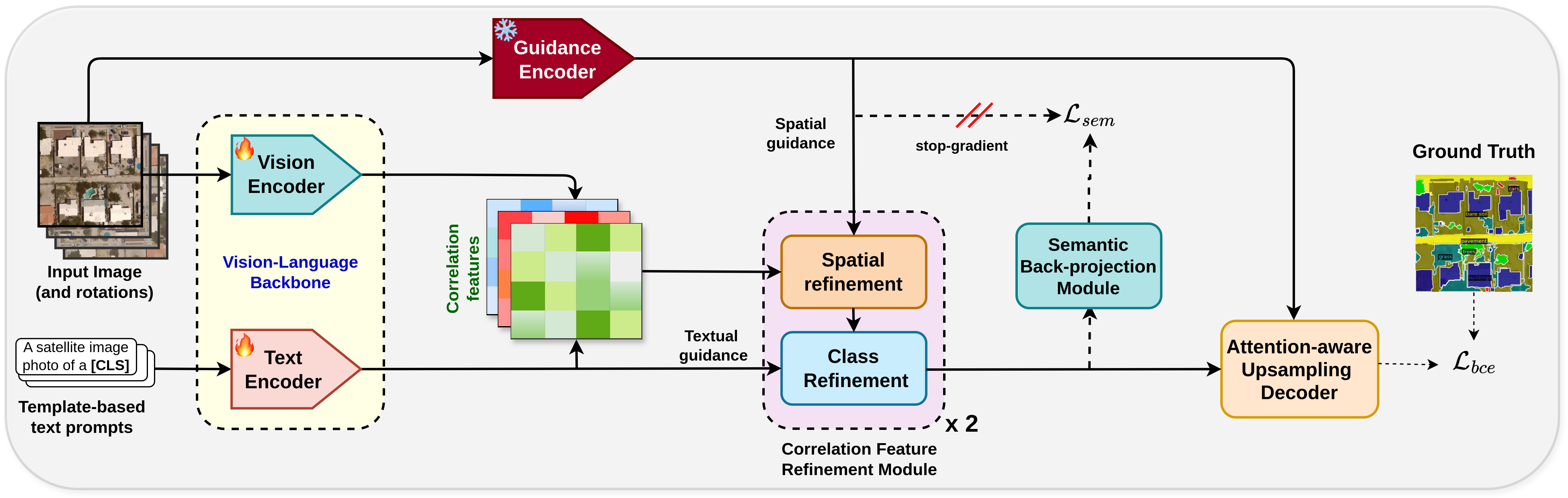}
    \caption{Overview of our proposed framework, \textbf{AerOSeg}. The input image (along with its rotated versions) and domain-specific text prompts are first processed by the Vision-Language backbone. The extracted image and text features are then used to generate Correlation features, which are refined through Correlation Feature Refinement blocks. The refined Correlation features are subsequently fed into Attention-aware Upsampling Decoder to yield final segmentation map. Additionally, features from the Guidance encoder are leveraged in both the Correlation Feature Refinement and  Attention-aware Upsampling Decoder. }
    \vspace{-15px}
    \label{fig:main}
\end{figure*}

In this work, we propose a novel segmentation framework, AerOSeg, composed of six interconnected modules: (i) Vision-Language Backbone, (ii) Guidance Encoder, (iii) Correlation Feature Computation, (iv) Correlation Feature Refinement, (v) Semantic Back-Projection Module, and (vi) Attention-Aware Upsampling Decoder. Figure~\ref{fig:main} provides an overview of the entire framework. In the following sections, we describe each component in detail and explain how they collectively contribute to our segmentation pipeline.

\subsection{Describing the Model Components}

\textbf{Vision-Language (VL) Backbone:}  
The VL backbone is a critical component of our framework, as it encapsulates the alignment between visual and linguistic feature spaces. In our approach, we employ CLIP ViT-B as the VL backbone. Given an input image \( I \), we first rotate it by a set of predefined angles and then pass both the original and rotated images through the CLIP vision encoder \( \mathcal{F}_v \) to extract dense image embeddings. These embeddings are subsequently rotated back to the original orientation. Specifically, for each rotation angle \( \theta \in \Theta = \{0^\circ, 90^\circ, 180^\circ, 270^\circ\} \), the dense embedding is computed as

\begin{equation}
    F_v^{\theta} = \operatorname{rotate}\big(\mathcal{F}_v(I^\theta), -\theta\big) \in \mathbb{R}^{(H \times W) \times d}.
    \label{eq:rotate_embedding}
\end{equation}

This multi-angle strategy enhances robustness to angular variations \cite{cao2024ovrs}. To generate textual embeddings, we create multiple text prompts for each candidate class \( c \) using remote-sensing-specific prompt templates, as suggested by Li et al. \cite{li2023rsclip}. In contrast to generic prompts (e.g., “A photo of a \texttt{[CLS]} in the scene” \cite{catseg,cao2024ovrs}), we employ the following templates:
\begin{itemize}[label=\ding{212}, leftmargin=2em]
    \item \texttt{A satellite image of a \texttt{[CLS]}} 
    \item \texttt{A land use image of a \texttt{[CLS]}} 
    \item \texttt{A remote sensing image of a \texttt{[CLS]}} 
    \item \texttt{An aerial image of a \texttt{[CLS]}} 
\end{itemize}

These prompts are processed by the CLIP text encoder \( \mathcal{F}_l \) to yield the textual embeddings:

\begin{equation}
    F_l = \mathcal{F}_l(T) \in \mathbb{R}^{N_C \times P \times d},
    \label{eq:text_embedding}
\end{equation}
where \( P = 4 \) denotes the number of prompts per class.

\textbf{Guidance Encoder:}  
Complementary to the VL backbone, the guidance encoder is designed to provide additional features that improve segmentation performance. Inspired by the success of SAM in diverse domains, we adopt its image encoder---a Vision Transformer (ViT)---as our guidance encoder. To capture a rich hierarchy of representations, we extract features from multiple stages of the encoder. In particular, we use the outputs from the 8\textsuperscript{th}, 16\textsuperscript{th}, and final layers of the SAM-L encoder, denoted as \( F_g^1 \), \( F_g^2 \), and \( F_g^3 \), respectively.

\textbf{Correlation Feature Computation:}  
With both image embeddings \( F_v^\theta(x) \) and text embeddings \( F_l(n,i) \) available (where \( x \) denotes a 2D spatial position, \( \theta \) the rotation angle, \( n \) the class index, and \( i \) the prompt index), we compute the image-text correlation map \( C^\theta \in \mathbb{R}^{(H \times W) \times N_C} \) using cosine similarity \cite{rocco2017convolutional}. For a given spatial location \( x \), class \( n \), and prompt \( i \), the correlation is defined as

\begin{equation}
    C^\theta(x,n,i) = \frac{F_v^\theta(x) \cdot F_l(n,i)}{\|F_v^\theta(x)\| \, \|F_l(n,i)\|}.
    \label{eq:cosine_similarity}
\end{equation}

Correlation maps computed across different rotation angles and prompts are concatenated and processed by a convolutional layer to generate the initial correlation feature \(\phi \in \mathbb{R}^{(H \times W) \times N_C \times d_\phi}\), where \( d_\phi \) denotes the feature dimension:

\begin{equation}
    \phi(x,n) = \operatorname{conv}\!\Biggl(\operatorname{concat}\!\Bigl(\Bigl\{ \bigl\{ C^\theta(x,n,i) \bigr\}_{i=1}^{P} \Bigr\}_{\theta \in \Theta}\Bigr)\!\Biggr).
    \label{eq:correlation_feature}
\end{equation}

\textbf{Correlation Feature Refinement Block:}  
Because CLIP is trained with a global contrastive objective, the dense features it produces can be noisy. To obtain reliable segmentation maps, the initial correlation features \( \phi \) must be refined. Following \cite{catseg}, our refinement module consists of two sequential submodules: (a) a Spatial Refinement Block and (b) a Class Refinement Block.

\begin{figure}[!t]
    \centering
    \setlength{\tabcolsep}{2pt} 
    \resizebox{0.4\textwidth}{!}{
    \begin{tabular}{ccc}
        {\includegraphics[width=0.15\textwidth]{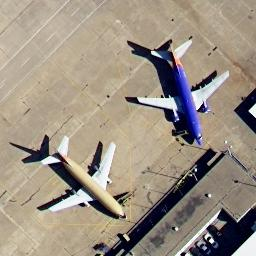}} & 
        {\includegraphics[width=0.15\textwidth]{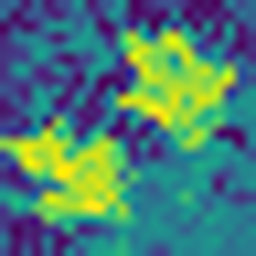}} & 
        {\includegraphics[width=0.15\textwidth]{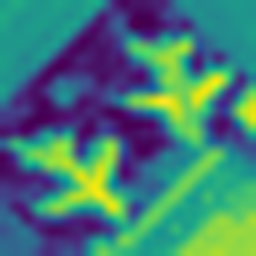}} \\

        \includegraphics[width=0.15\textwidth]{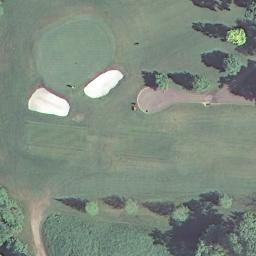} &
        \includegraphics[width=0.15\textwidth]{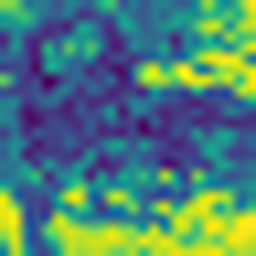} &
        \includegraphics[width=0.15\textwidth]{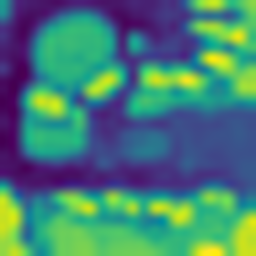} \\

        \subcaptionbox{Input image}{\includegraphics[width=0.15\textwidth]{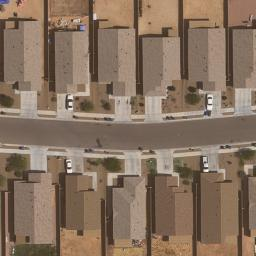}} & 
        
        \subcaptionbox{CLIP features}{\includegraphics[width=0.15\textwidth]{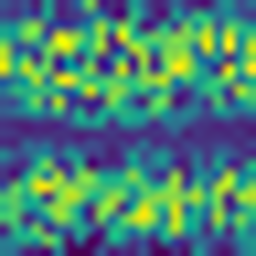}} & 
        \subcaptionbox{SAM features}{\includegraphics[width=0.15\textwidth]{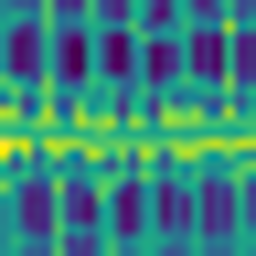}} \\

    \end{tabular}
    }
    \caption{\textbf{Feature visualization for CLIP and SAM features.} SAM produces feature maps with richer semantic information and reduced noise compared to CLIP. }
    \vspace{-15px}
    \label{fig:viz1}
\end{figure}

\textit{(a) Spatial Refinement Block:}  
To improve the spatial structure of the correlation features, we apply a Swin Transformer~\cite{liu2021swin} module. Refinement is performed independently for each class using two successive Swin Transformer blocks: the first applies window-based multi-head self-attention (W-MSA) over local windows, and the second uses shifted window-based self-attention (SW-MSA). Additionally, to further enhance spatial refinement, we integrate guidance from dense visual features.
While prior work \cite{catseg,cao2024ovrs} utilizes CLIP visual embeddings for this task, it has several drawbacks. First, since correlation features are originally derived from interactions between CLIP image and text embeddings, using CLIP features for refinement is suboptimal. Second, dense features from the CLIP image encoder tend to be noisy due to its global contrastive learning objective. In contrast, SAM produces more semantic and less noisy feature maps (see Fig.~\ref{fig:viz1}). Therefore, we leverage SAM-derived features in the spatial refinement block. The refined feature $\phi'$ is computed as:

\begin{equation}
    \phi'(:, n) = \mathcal{T}^{\mathrm{sp}}\Bigl(\phi(:, n),\, \mathcal{P}_v(F_g^3)\Bigr),
    \label{eq:spatial_refinement}
\end{equation}

where \(\mathcal{T}^{\mathrm{sp}}\) denotes the Spatial Refinement Block and \(\mathcal{P}_v\) is a projection layer. The correlation features and visual embeddings are concatenated to form the query and key, while only the correlation features are used as the value feature in this block. 
Please refer to Sec. 2 of  Supplementary Material for visualizations of the effect of using SAM features in the Spatial Refinement Block.

\textit{(b) Class Refinement Block:}  
After spatial refinement, we further process the features to incorporate the text modality and explicitly capture inter-class relationships. This step is crucial for addressing the challenges of open-vocabulary segmentation, such as handling a variable number of categories and ensuring invariance to their order. To meet these requirements, we employ a linear transformer layer without positional embeddings \cite{katharopoulos2020transformers}. In this block, the text embeddings \( F_l \) serve as guidance features, and the refined output is given by

\begin{equation}
    \phi''(x,:) = \mathcal{T}^{\mathrm{cls}}\Bigl(\phi'(x,:),\, \mathcal{P}_l(\bar{F}_l)\Bigr),
    \label{eq:class_refinement}
\end{equation}

where \(\mathcal{T}^{\mathrm{cls}}\) is the Class Refinement Block, \(\mathcal{P}_l\) a projection layer, and \(\bar{F}_l\) represents the text embeddings averaged over all prompts. In our work, two Correlation Feature Refinement blocks are used.

\begin{figure}[!t]
    \centering
    \includegraphics[width=\linewidth]{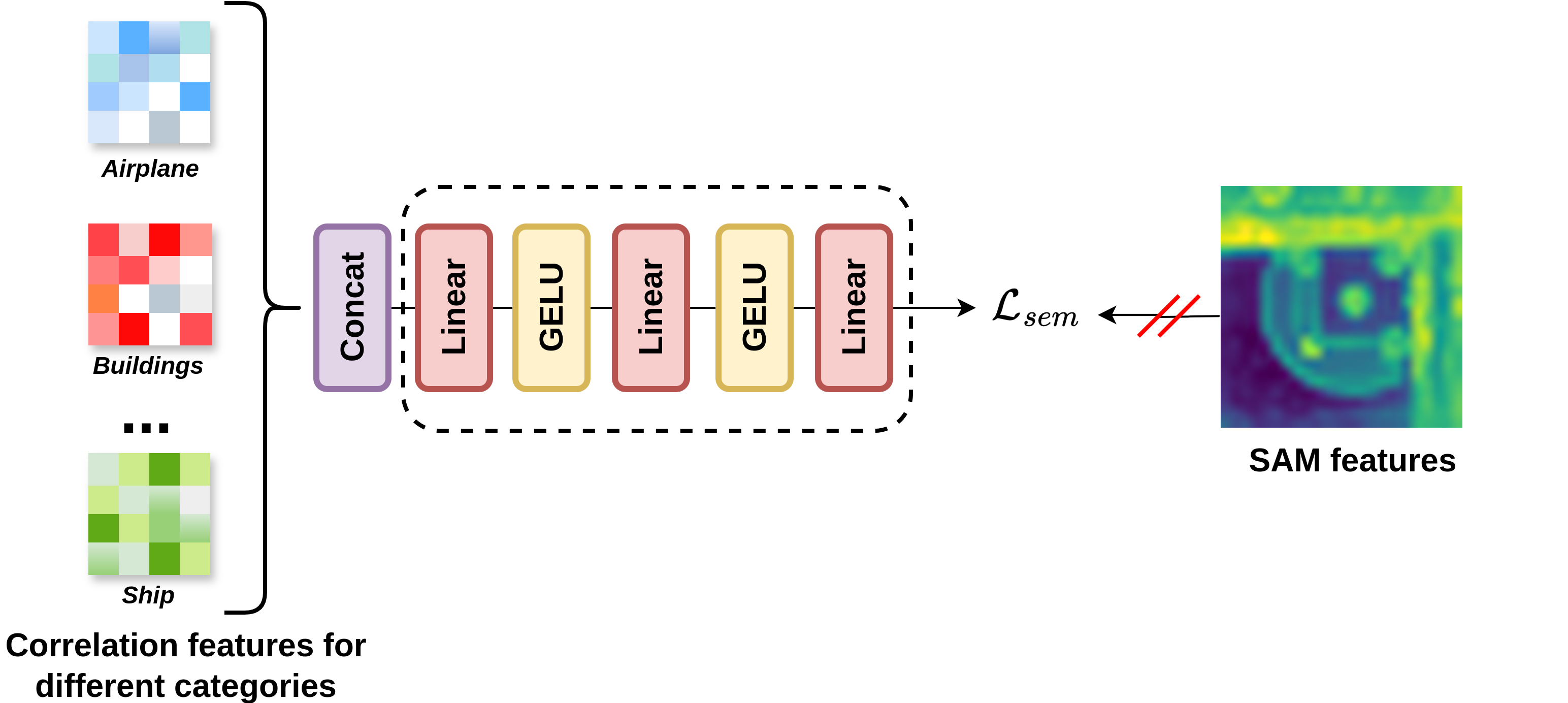}
    \caption{\textbf{Semantic Back-projection module and loss.} Class-specific correlation maps are concatenated and passed to an MLP block. Semantic back-projection loss $\mathcal{L}_{sem}$ is computed between reconstructed features and SAM features.}
    \vspace{-15px}
    \label{fig:sbp}
\end{figure}

\textbf{Semantic Back-Projection Module:}  
To ensure that the refined correlation features remain semantically consistent with SAM features, we introduce a lightweight back-projection module. This module reconstructs the SAM features from the class-wise refined correlation features \( \phi'' \). Specifically, the correlation features are concatenated along the channel dimension and processed through three linear layers with GELU activation, yielding the reconstructed feature \( \psi \in \mathbb{R}^{(H \times W) \times d} \) as follows:

\begin{equation}
    \psi = \mathcal{T}^{sem}\!\Biggl(\operatorname{concat}\!\Bigl(\Bigl\{  \phi''(:,n) \Bigr\}_{n=1}^{N_\mathcal{C}}\Bigr)\!\Biggr).
\end{equation}

where, $\mathcal{T}^{sem}$ is Semantic back-projection module and $N_\mathcal{C}$ is the number of classes. The reconstructed feature is subsequently used to compute a Semantic Back-Projection Loss (see Section~\ref{subsec:loss}). Fig.~\ref{fig:sbp} shows a schematic of Semantic back-projection module. Our design is inspired by the Recovery Decoder in \cite{clip_rc}, yet differs in that we reconstruct SAM features using a lightweight MLP, rather than reconstructing CLIP features via multi-head cross-attention.

\textbf{Attention-aware Upsampling Decoder:}  
Finally, as the refined correlation features \( \phi'' \) are at 1/16th of the input image resolution, an attention-aware upsampling decoder is employed to restore full resolution. Initially, \( \phi'' \) is upsampled by a factor of 2 using a transposed convolution, yielding \( \phi'_{\text{2x}} \). Spatial and channel attention features \cite{cao2024ovrs} are then computed from the class-averaged upsampled features \( \bar{\phi}'_{\text{2x}} \) as follows:
\vspace{-10px}
\begin{align}
    A^{sp} &= \operatorname{conv}\Bigl(\operatorname{avgpool}_{sp}\bigl(\bar{\phi}'_{\text{2x}}\bigr)\Bigr), \label{eq:spatial_attention}\\[1mm]
    A^{ch} &= \operatorname{conv}\Bigl(\operatorname{avgpool}_{ch}\bigl(\bar{\phi}'_{\text{2x}}\bigr)\Bigr), \label{eq:channel_attention}
\end{align}

where \(\operatorname{avgpool}_{sp}\) and \(\operatorname{avgpool}_{ch}\) denote average pooling across the spatial and channel dimensions, respectively. The intermediate guidance features \( F^2_g \) from SAM are then transformed via
\vspace{-5px}
\begin{equation}
    F'_g = A^{sp} \odot \operatorname{up}(F^2_g, 2) + A^{ch} \odot \operatorname{up}(F^2_g, 2) + \operatorname{up}(F^2_g, 2),
    \label{eq:guidance_transform}
\end{equation}
where \(\odot\) represents the Hadamard product and \(\operatorname{up}(\cdot,2)\) denotes nearest neighbor upsampling by a factor of 2. The upsampled correlation feature \( \phi_{\text{2x}} \) is obtained by concatenating \( \phi'_{\text{2x}} \) with the transformed guidance feature \( F'_g \) and applying a convolution:

\begin{equation}
    \phi_{\text{2x}} = \operatorname{conv}\Bigl([\phi'_{\text{2x}},\, F'_g]\Bigr).
    \label{eq:upsampled_feature}
\end{equation}

This upsampling process is repeated to generate \( \phi_{\text{4x}} \), which is then refined via a convolution layer and bilinear upsampling to yield the final segmentation map \( \hat{y} \).

\begin{table*}[!h]
\centering
\begin{center} 

\renewcommand{\arraystretch}{1.25}
\resizebox{1\textwidth}{!}{%
\begin{tabular}{
  c|c
}
\hline
\textbf{Dataset} & \textbf{Classes}\\
\hline
\multirow{2}{*}{\textbf{iSAID}}   & \textcolor{deepgreen}{ship, storage tank, baseball diamond, basketball court, ground track field, large vehicle, swimming pool, roundabout, plane,}\\

\cdashline{2-2}[1.5pt/2pt] 
 & \textcolor{magenta}{tennis court, bridge, small vehicle, helicopter, soccer ball field, harbor}                                                    \\
\hline
\multirow{2}{*}{\textbf{DLRSD}}   & \textcolor{deepgreen}{chaparral, court, dock, field, grass, mobile home, sand, ship, tanks, water,} \\
\cdashline{2-2}[1.5pt/2pt] 
& \textcolor{magenta}{airplane, bare soil, buildings, cars, pavement, sea, trees}                                                                                                                   \\
\hline
\multirow{2}{*}{\textbf{OEM}}    & \textcolor{deepgreen}{bareland, rangeland, road, building,} \\
\cdashline{2-2}[1.5pt/2pt] 
& \textcolor{magenta}{developed space, tree, water, agriculture land} \\
\hline
\end{tabular}
}
\end{center}
\vspace{-10px}
\caption{Class details for different datasets used in our work. Classes in \textcolor{deepgreen}{Green} are seen during training, whereas classes in \textcolor{magenta}{Magenta} are only encountered in inference.}
\vspace{-10px}
\label{tab:dataset}
\end{table*}

\subsection{Loss Functions}
\label{subsec:loss}

To train our network, we utilize a combination of loss functions that jointly optimize segmentation accuracy and feature semantic consistency.

\textbf{Binary Cross-Entropy Loss:}  
We use the binary cross-entropy loss to align the predicted segmentation maps with the ground truth. The loss is defined as
\vspace{-5px}
\begin{equation}
    \mathcal{L}_{bce} = \frac{1}{HW} \sum_{c \in \mathcal{C}} \sum_{i,j} \Bigl[ -y^c_{ij}\log\bigl(\hat{y}^c_{ij}\bigr) - \bigl(1-y^c_{ij}\bigr)\log\bigl(1-\hat{y}^c_{ij}\bigr) \Bigr],
    \label{eq:bce_loss}
\end{equation}

where \( \hat{y}^c_{ij} \) and \( y^c_{ij} \) denote the predicted probability and ground truth for class \( c \) at pixel \((i,j)\), respectively.

\textbf{Semantic Back-Projection Loss:}  
To ensure that the reconstructed feature \( \psi \) is semantically consistent with the SAM features from the guidance encoder, we introduce the semantic back-projection loss:
\vspace{-5px}
\begin{equation}
    \mathcal{L}_{sem} = \|\psi - sg(F_g^3)\|_2^2.
    \label{eq:sem_loss}
\end{equation}
where $sg(.)$ is stop-gradient operator.

\textbf{Overall Loss:}  
The final loss function is a linear combination of the binary cross-entropy loss and the semantic back-projection loss:
\vspace{-5px}
\begin{equation}
    \mathcal{L} = \mathcal{L}_{bce} + \mathcal{L}_{sem}.
    \label{eq:total_loss}
\end{equation}

\vspace{-10px}

\section{Experiments}

\subsection{Dataset details}

\noindent \textbf{iSAID} dataset is originally an instance segmentation dataset developed from large-scale object detection dataset DOTA \cite{8578516}, which consists of aerial images collected from multiple sensors. The dataset consists of varying high-resolution images and their annotations for 15 object categories. We utilize the processed iSAID dataset \cite{cao2024ovrs,yao2021scale} which consists of 18,076 training and 6,363 validation images of $256 \times 256$ resolution.

\noindent \textbf{DLRSD} - Dense Labeling Remote Sensing Dataset is an extension of the multi-label Remote Sensing Image Retrieval (RSIR) archive \cite{dlrsd}. The images from this archive were semantically divided and assigned predefined pixel labels to enable various downstream tasks beyond image retrieval. The processed DLRSD dataset \cite{rs10060964} consists of 7002 aerial images with a spatial resolution of $256 \times 256$, along with annotations for 17 object categories. The training set contains 5601 images, while the validation set includes 1401 images.

\noindent \textbf{OpenEarthMap} \cite{xia_2023_openearthmap} (OEM) is a global high-resolution land cover mapping dataset comprising 5,000 aerial and satellite images with annotations for eight categories. It is constructed using images from existing benchmark datasets, covering 97 regions across 44 countries, with a ground sampling distance of 0.25–0.5m. For our experiments, we excluded the xBD subset due to partial availability. Our processed dataset contains 31,154 training images and 5,195 validation images, each of size $256 \times 256$.

For each dataset, we split the classes into seen and unseen categories. Only the seen classes are available during training, while all classes are used during inference. Details of the seen-unseen class split are provided in Table~\ref{tab:dataset}.

\subsection{Implementation details}

We develop our models using PyTorch and Detectron2 framework. We have used initial learning rate of $2 \times 10^{-6}$ for VL backbone. Following \cite{catseg}, we only fine-tune query and value projection matrices of image and text encoders in the VL backbone keeping other parameters frozen. Initial learning rate is set to $2 \times 10^{-4}$ for every other module except guidance encoder. We have kept parameters of guidance encoder frozen throughout training. We have used AdamW optimizer with a batch size of 4 to train our models. Our model is trained for 10K, 5K and 15K iterations for iSAID, DLRSD and OEM datasets respectively. We use one NVIDIA A100 80GB GPU to run our experiments.

\begin{table*}[!htp]
\centering
\renewcommand{\arraystretch}{1.25}
\resizebox{1\textwidth}{!}{%
\begin{tabular}{c|c|c c c|c c c|c c c : c c c}
\hline
\multirow{2}{*}{\textbf{Method}} &\multirow{2}{*}{\textbf{Venue}} & \multicolumn{3}{c}{\textbf{iSAID}} & \multicolumn{3}{c}{\textbf{DLRSD}}& \multicolumn{3}{c}{\textbf{OEM}} & \multicolumn{3}{c}{\textbf{Average}} \\ 
\cline{3-14}
 & & s-mIoU & u-mIoU & h-mIoU & s-mIoU & u-mIoU & h-mIoU & s-mIoU & u-mIoU & h-mIoU & s-mIoU & u-mIoU & h-mIoU \\
\hline
SAN & CVPR'23 & 62.28 & 35.77 & 45.44 & 50.18 & 12.62 & 20.17 & 45.36 & 15.19 & 22.76 & 52.61 & 21.19 & 29.46 \\
SCAN & CVPR'24 &44.31 &35.55 &39.45 &25.86 &20.10 & 22.62& 31.85& 16.53&21.77 &34.01 & 24.06&27.95 \\
SED & CVPR'24 & 74.37 & 36.00 & 48.51 & \textcolor{red}{77.73} & 27.00 & 40.08 & \textcolor{red}{57.46} & 46.95 & \textcolor{red}{51.68} & \textcolor{red}{69.85} & 36.65 & 46.76 \\
CAT-Seg & CVPR'24 & 72.57 & 42.64 & 53.70 & 52.79 & 24.39 & 33.36 & 49.33 & 39.51 & 43.87 & 58.23 & 35.51 & 43.64 \\
OVRS & arXiv'24 & \textcolor{red}{75.85} & \textcolor{blue}{49.27} & \textcolor{blue}{59.73} & \textcolor{blue}{61.00} & \textcolor{blue}{31.86} & \textcolor{blue}{41.86} & 51.20 &\textcolor{blue}{ 47.23 }& 49.13 & \textcolor{blue}{62.68} & \textcolor{blue}{42.79} & \textcolor{blue}{50.24} \\
\rowcolor{gray!30}\textbf{\ourmodel\ (Ours)} & - & \textcolor{blue}{75.48} & \textcolor{red}{51.46} & \textcolor{red}{61.20} & 60.37 & \textcolor{red}{39.12} & \textcolor{red}{47.48} & \textcolor{blue}{51.27} & \textcolor{red}{48.16} & \textcolor{blue}{49.66} & 62.37 & \textcolor{red}{46.25} & \textcolor{red}{52.78} \\
\hline
\end{tabular}
}
\caption{\textbf{Quantitative comparison with state-of-the-art methods.} Values in \textcolor{red}{red} and \textcolor{blue}{blue} indicate the best and second-best results, respectively. The results show that while other methods tend to overfit seen classes, our approach generalizes well to both seen and unseen classes, achieving the highest average h-mIoU score.}
\vspace{-10px}
\label{tab:sota}
\end{table*}

\begin{figure*}[!h]
    \centering
    \renewcommand{\arraystretch}{-5.0} 
    \setlength{\tabcolsep}{-3pt} 

    \resizebox{0.95\textwidth}{!}{
    \begin{tabular}{@{}cccccc@{}}
    
        \begin{tikzpicture}
            \node at (0,0) {\includegraphics[width=0.16\textwidth]{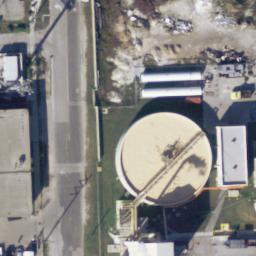}};
            \draw[red, thick, densely dashed] (-1.3,-1) rectangle (0.0,1.25);
        \end{tikzpicture} &
        \begin{tikzpicture}
            \node at (0,0) {\includegraphics[width=0.16\textwidth]{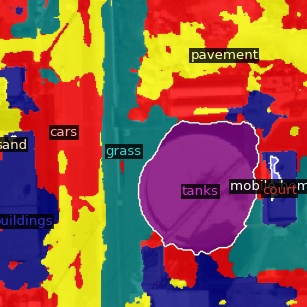}};
            \draw[red, thick, densely dashed] (-1.3,-1) rectangle (0.0,1.25);
        \end{tikzpicture} &
        \begin{tikzpicture}
            \node at (0,0) {\includegraphics[width=0.16\textwidth]{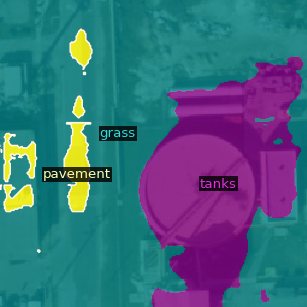}};
            \draw[red, thick, densely dashed] (-1.3,-1) rectangle (0.0,1.25);
        \end{tikzpicture} &
        \begin{tikzpicture}
            \node at (0,0) {\includegraphics[width=0.16\textwidth]{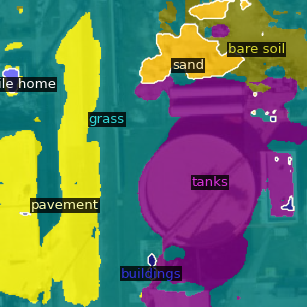}};
            \draw[red, thick, densely dashed] (-1.3,-1) rectangle (0.0,1.25);
        \end{tikzpicture} &
        \begin{tikzpicture}
            \node at (0,0) {\includegraphics[width=0.16\textwidth]{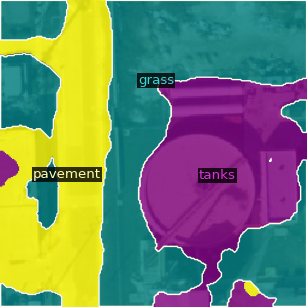}};
            \draw[red, thick, densely dashed] (-1.3,-1) rectangle (0.0,1.25);
        \end{tikzpicture} &
        \begin{tikzpicture}
            \node at (0,0) {\includegraphics[width=0.16\textwidth]{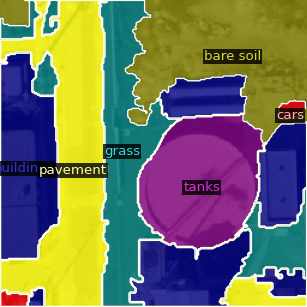}};
            \draw[red, thick, densely dashed] (-1.3,-1) rectangle (0.0,1.25);
        \end{tikzpicture} \\

        \begin{tikzpicture}
            \node at (0,0) {\includegraphics[width=0.16\textwidth]{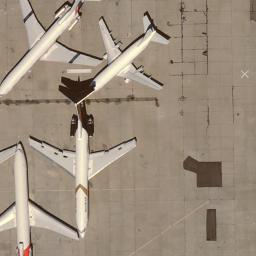}};
            \draw[red, thick, densely dashed] (-1.3,-0.3) rectangle (0.7,1.25);
        \end{tikzpicture} &
        \begin{tikzpicture}
            \node at (0,0) {\includegraphics[width=0.16\textwidth]{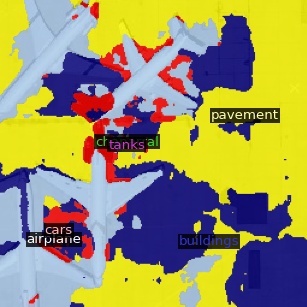}};
            \draw[red, thick, densely dashed] (-1.3,-0.3) rectangle (0.7,1.25);
        \end{tikzpicture} &
        \begin{tikzpicture}
            \node at (0,0) {\includegraphics[width=0.16\textwidth]{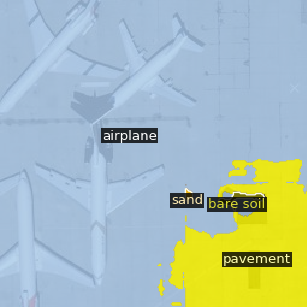}};
            \draw[red, thick, densely dashed] (-1.3,-0.3) rectangle (0.7,1.25);
        \end{tikzpicture} &
        \begin{tikzpicture}
            \node at (0,0) {\includegraphics[width=0.16\textwidth]{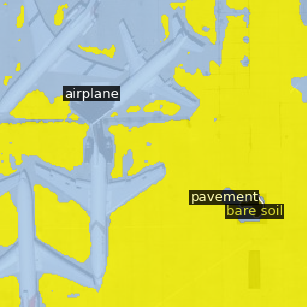}};
            \draw[red, thick, densely dashed] (-1.3,-0.3) rectangle (0.7,1.25);
        \end{tikzpicture} &
        \begin{tikzpicture}
            \node at (0,0) {\includegraphics[width=0.16\textwidth]{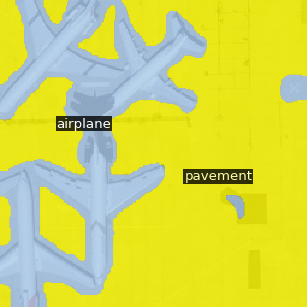}};
            \draw[red, thick, densely dashed] (-1.3,-0.3) rectangle (0.7,1.25);
        \end{tikzpicture} &
        \begin{tikzpicture}
            \node at (0,0) {\includegraphics[width=0.16\textwidth]{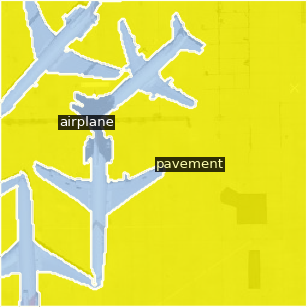}};
            \draw[red, thick, densely dashed] (-1.3,-0.3) rectangle (0.7,1.25);
        \end{tikzpicture} \\

        \subcaptionbox{Input image}{
        \begin{tikzpicture}
            \node at (0,0) {\includegraphics[width=0.16\textwidth]{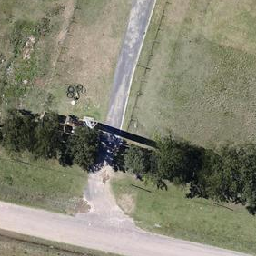}};
            \draw[red, thick, densely dashed] (0,0) rectangle (1.25,1.25);
            \draw[orange, thick, densely dashed] (-1.25,-0.7) rectangle (-0.1,0.35);
        \end{tikzpicture}} &
        \subcaptionbox{SED}{
        \begin{tikzpicture}
            \node at (0,0) {\includegraphics[width=0.16\textwidth]{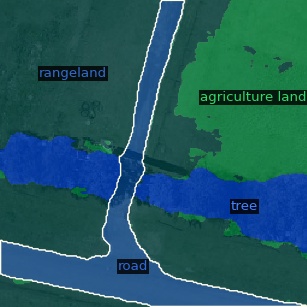}};
            \draw[red, thick, densely dashed] (0,0) rectangle (1.25,1.25);
            \draw[orange, thick, densely dashed] (-1.25,-0.7) rectangle (-0.1,0.35);
        \end{tikzpicture}} &
        \subcaptionbox{CAT-Seg}{
        \begin{tikzpicture}
            \node at (0,0) {\includegraphics[width=0.16\textwidth]{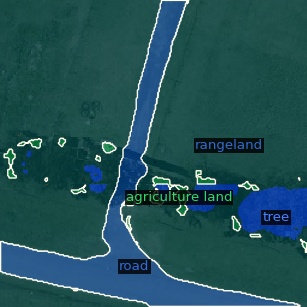}};
            \draw[red, thick, densely dashed] (0,0) rectangle (1.25,1.25);
            \draw[orange, thick, densely dashed] (-1.25,-0.7) rectangle (-0.1,0.35);
        \end{tikzpicture}} &
        \subcaptionbox{OVRS}{
        \begin{tikzpicture}
            \node at (0,0) {\includegraphics[width=0.16\textwidth]{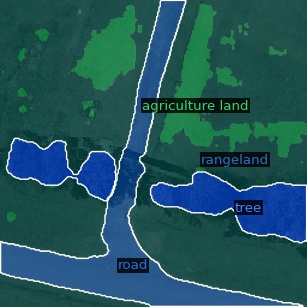}};
            \draw[red, thick, densely dashed] (0,0) rectangle (1.25,1.25);
            \draw[orange, thick, densely dashed] (-1.25,-0.7) rectangle (-0.1,0.35);
        \end{tikzpicture}} &
        \subcaptionbox{AerOSeg (Ours)}{
        \begin{tikzpicture}
            \node at (0,0) {\includegraphics[width=0.16\textwidth]{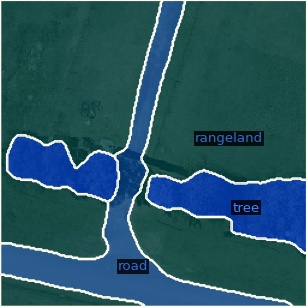}};
            \draw[red, thick, densely dashed] (0,0) rectangle (1.25,1.25);
            \draw[orange, thick, densely dashed] (-1.25,-0.7) rectangle (-0.1,0.35);
        \end{tikzpicture}} &
        \subcaptionbox{Ground Truth}{
        \begin{tikzpicture}
            \node at (0,0) {\includegraphics[width=0.16\textwidth]{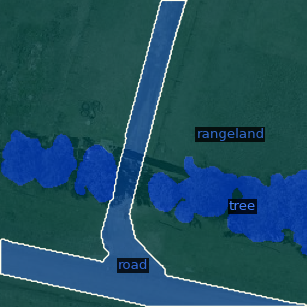}};
            \draw[red, thick, densely dashed] (0,0) rectangle (1.25,1.25);
            \draw[orange, thick, densely dashed] (-1.25,-0.7) rectangle (-0.1,0.35);
        \end{tikzpicture}} \\

    \end{tabular}
    }
    \vspace{-5px}
    \caption{\textbf{Qualitative comparison with state-of-the-art methods.} Dashed bounding boxes highlight regions where our model achieves more precise segmentation with sharper boundaries.}
    \vspace{-15px}
    \label{qual_comp}
\end{figure*}
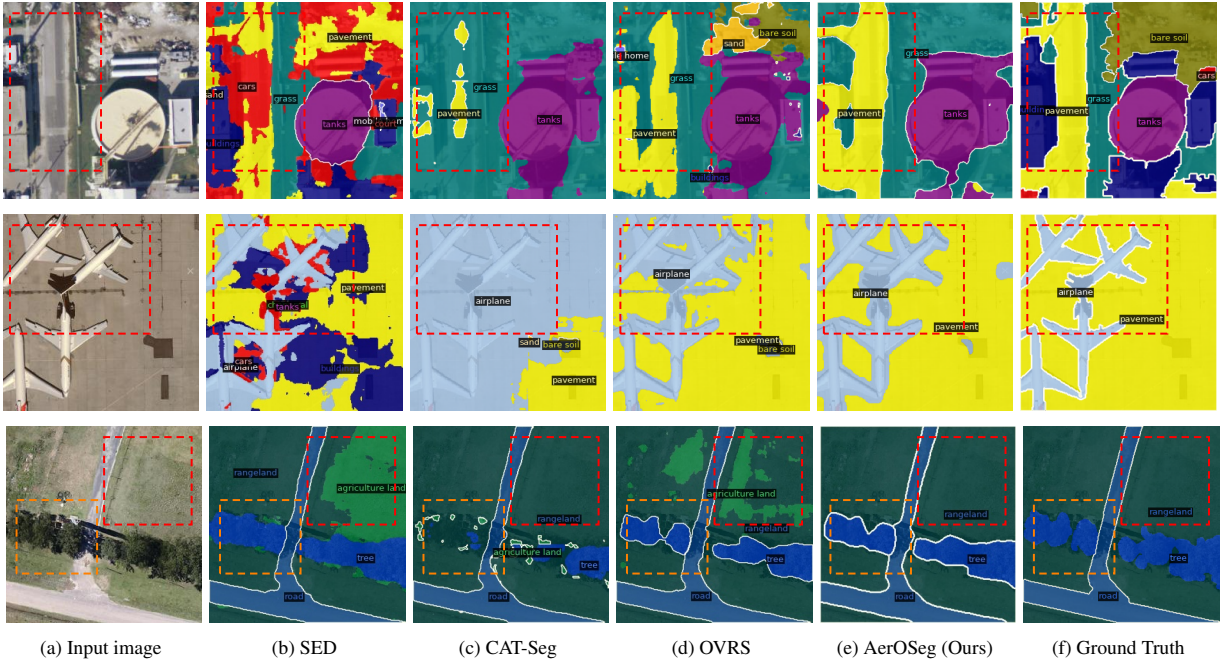

\subsection{Evaluation metrics}

We compute Intersection-over-Union (IoU) metric for all the classes. We report mean IoU score over seen classes and unseen classes separately, denoted as \textbf{s-mIoU} and \textbf{u-mIoU} respectively as used commonly in Open-Vocabulary Segmentation literature \cite{zhou2023zegclip, zegformer}. Additionally, we report the harmonic mean of s-mIoU and u-mIoU, denoted as \textbf{h-mIoU}, which serves as the key metric for evaluating generalized zero-shot performance.

\subsection{Comparison with state-of-the-art}

We compare our approach with several state-of-the-art Open-Vocabulary Segmentation methods: SAN \cite{san}, SCAN \cite{liu2024scan}, SED \cite{xie2024sed}, CAT-Seg \cite{catseg}, and OVRS \cite{cao2024ovrs}. Among these, OVRS is designed for remote sensing, while the others are originally developed for natural images. These models are trained using their publicly available source codes with default configurations. We use their base versions, where SED employs a CLIP ConvNeXT-B architecture, while the others utilize a CLIP ViT-B/16 as the Vision-Language backbone. Table~\ref{tab:sota} presents a quantitative comparison of these methods.
\ourmodel\ achieves higher h-mIoU scores than other state-of-the-art methods on the iSAID and DLRSD datasets and ranks second on the OEM dataset. On average, \ourmodel\ surpasses the next best model, OVRS, by a margin of 2.54 h-mIoU, highlighting its strong generalization capability.

In Fig. \ref{qual_comp}, we qualitatively compare segmentation predictions across models. Our model achieves more precise segmentation for ``Pavement'' in the first image and sharper boundaries for ``Airplane'' in the second. In the third row, it accurately segments ``Tree'' and ``Rangeland". 

For comparison with training-free method SegEarth-OV \cite{li2024segearth} and more qualitative results, please refer to Sec. 3 of Supplementary material.

\vspace{-5px}
\subsection{Ablation study}
\vspace{-5px}
\begin{table*}[!h]
\centering
\renewcommand{\arraystretch}{1.25}
\resizebox{0.75\textwidth}{!}{
\begin{tabular}{
  c>{\columncolor{yellow!15}}c>{\columncolor{green!15}}c>{\columncolor{red!15}}cccc:c
}
\hline
\textbf{Configuration} & \makecell{\textbf{Guidance}\\\textbf{Encoder}} & \makecell{\textbf{Semantic Back-}\\\textbf{projection loss}} & \makecell{\textbf{Remote-Sensing}\\ \textbf{Prompts}} & \textbf{iSAID} & \textbf{DLRSD} & \textbf{OEM} & \textbf{Average} \\
\hline
Baseline      & \ding{55}           & -~                               & \ding{55}                 & 57.88  & 28.69  & 46.86  &  44.47 \\
Config-A      & \ding{51}           & \ding{55}                        & \ding{55}                 & 58.10  & 39.93  &   \textbf{50.46}     &  49.49 \\
Config-B      & \ding{51}           & \ding{51}                        & \ding{55}                 & 59.32  & 42.34  & 47.16  &  49.60 \\
\cellcolor{gray!20}Config-C \textbf{(Ours)}     & \ding{51}  & \ding{51} & \ding{51} & \cellcolor{gray!20}\textbf{61.20} & \cellcolor{gray!20}\textbf{47.48} & \cellcolor{gray!20}49.66 & \cellcolor{gray!20}\textbf{52.78} \\
\hline
\end{tabular}
}
\vspace{-5px}
\caption{Ablation study for different components of our framework.}
\vspace{-15px}
\label{tab:abl_1}
\end{table*}

\hspace{1em} \textbf{Main components ablation:} We experimented with multiple configurations to evaluate the effectiveness of different components in our framework:  

\begin{itemize}
    \item \textbf{(I) Baseline:} Our model without the Guidance Encoder, trained only with $\mathcal{L}_{bce}$. A single generic prompt, \texttt{"A photo of a [CLS] in a scene"}, is used instead of the Remote-Sensing prompt ensemble. Intermediate CLIP features serve as guidance for spatial refinement and the decoder.  
    \item \textbf{(II) Config-A:} Our model with the Guidance Encoder and a single generic prompt, trained with $\mathcal{L}_{bce}$.  
    \item \textbf{(III) Config-B:} Our model with the Guidance Encoder and a single generic prompt, trained with $\mathcal{L}_{bce}$ and $\mathcal{L}_{sem}$.  
    \item \textbf{(IV) Config-C:} Our full model.  
\end{itemize}

Table~\ref{tab:abl_1} presents the h-mIoU scores for these configurations. Adding the Guidance Encoder in Config-A improves the average h-mIoU by 5.02 over the Baseline. Training with Semantic Back-Projection loss further enhances h-mIoU by 1.22 and 2.41 for iSAID and DLRSD, respectively. Finally, incorporating the Remote Sensing prompt ensemble yields additional gains of 1.88, 5.14, and 2.5 for the iSAID, DLRSD, and OEM datasets, respectively, compared to Config-B.

\textbf{Guidance encoder training: } We experimented with training the Guidance Encoder jointly with the rest of the network, fine-tuning it with an initial learning rate of $2 \times 10^{-6}$. A quantitative comparison of h-mIoU scores obtained for the fine-tuned and frozen Guidance Encoder is shown in Table~\ref{guidance_training}. While fine-tuning improves iSAID performance by 0.36 h-mIoU, the frozen Guidance Encoder yields significantly better results for DLRSD and OEM, with gains of 3.07 and 2.63 h-mIoU, respectively. We hypothesize that, given the relatively small dataset size, the frozen Guidance Encoder performs better on average.

\vspace{-5px}
\begin{table}[!h]
\centering
\renewcommand{\arraystretch}{1.3}
\resizebox{\linewidth}{!}{
\begin{tabular}{c|ccc:c}
\hline
\makecell{\textbf{Guidance Encoder}\\\textbf{Training Strategy}} & \textbf{iSAID}           & \textbf{DLRSD}                   & \textbf{OEM}           & \textbf{Average} \\
\hline
Fine-tune  \color{orange!80}\faFire             & ~\textbf{61.56} & 44.41                   & 47.03     & 51.00 \\
Frozen \textbf{(Ours)} \color{blue!40}\faSnowflake                      & 61.20           & \textbf{47.48} & \textbf{49.66} & \textbf{52.78} \\
\hline
\end{tabular}
}
\vspace{-5px}
\caption{Quantitative comparison between fine-tuned and frozen Guidance Encoder.}
\vspace{-10px}

\label{guidance_training}

\end{table}

\textbf{Ablation on Semantic Back-Projection Loss:} We experimented with different reconstruction targets for our proposed semantic back-projection module: (i) RGB, (ii) CLIP features, and (iii) SAM features. Table~\ref{tab:recon} compares the h-mIoU scores on three datasets for different reconstruction targets. Using RGB as the reconstruction target leads to significantly lower performance due to overfitting on known classes. Reconstructing CLIP features improves performance for both seen and unseen categories compared to RGB. However, the best results are achieved when SAM features are used as the reconstruction target.

\begin{table}[!h]
\centering
\renewcommand{\arraystretch}{1.3}
\resizebox{\linewidth}{!}{
\begin{tabular}{c|ccc:c}
\hline
\textbf{Reconstruction target} & \textbf{iSAID}         & \textbf{DLRSD}         & \textbf{OEM}         & \textbf{Average} 
\\
\hline
RGB                       & 14.93          & 40.03       & 38.21       & 31.06 \\
CLIP features             & 53.27          & 42.10       & 48.55      & 47.97 \\
SAM features \textbf{(Ours)}             & \textbf{61.20} & \textbf{47.48} & \textbf{49.66} & \textbf{52.78} \\
\hline
\end{tabular}
}
\vspace{-5px}
\caption{Ablation study on different reconstruction targets for the Semantic Back-Projection module.}
\vspace{-10px}
\label{tab:recon}
\end{table}

\textbf{Effect of Different Guidance Encoders:} We investigated various SAM image encoders as our guidance encoder, specifically experimenting with SAM, SAM2 \cite{ravi2024sam2}, and SAM2.1 \cite{ravi2024sam2}. While SAM is originally trained for promptable image segmentation, SAM2 and SAM2.1 are designed for promptable video segmentation. For SAM, we used the ViT-L and ViT-H variants, whereas for SAM2 and SAM2.1, we employed the Hiera-Base+ variant. Table-\ref{abl_sam} presents the quantitative results on the iSAID dataset using different SAM variants as the guidance encoder. Among these, SAM2.1 achieved the best overall performance.

\begin{table}[!h]
\centering
\resizebox{0.45\textwidth}{!}{
\begin{tblr}{
  cells = {c},
  hline{1,6} = {-}{0.08em},
  hline{2} = {-}{},
}
\textbf{Guidance Encoder} & \textbf{Architecture} & \textbf{s-mIoU} & \textbf{u-mIoU} & \textbf{h-mIoU} \\
SAM             & ViT-L        & 75.48  & 51.46  & 61.20  \\
SAM             & ViT-H        & 74.90  & 52.81  & 61.94  \\
SAM2              &     Hiera-Base+         &   \textcolor{red}{77.63}     &    \textcolor{blue}{53.32}    &  \textcolor{blue}{63.22}     \\
SAM2.1            &       Hiera-Base+       &   \textcolor{blue}{77.06}     &    \textcolor{red}{53.93}    &    \textcolor{red}{63.45}    
\end{tblr}
}
\caption{Quantitative comparison between different Guidance encoders on iSAID dataset.}
\vspace{-10px}
\label{abl_sam}
\end{table}

We have also explored the effect of different VL backbones, please refer to Sec. 1 of Supplementary Material for more details.
\vspace{-5px}
\section{Conclusion}
\vspace{-5px}
In this work, we introduce AerOSeg, a novel open-vocabulary segmentation framework for remote sensing that combines CLIP-based vision-language alignment with SAM-guided feature refinement. Our approach mitigates the semantic knowledge gap in CLIP by leveraging SAM as a Guidance Encoder. Additionally, Semantic Back-Projection loss ensures better alignment between image-text correlation features and SAM features, preserving semantic consistency. Extensive experiments on benchmark remote sensing datasets demonstrate that our model outperforms existing methods, excelling in generalization to both seen and novel classes. This work paves the way for advancing Open-Vocabulary Semantic Segmentation in geospatial applications.

\textbf{Acknowledgements.} We would like to thank C-MInDS, IIT Bombay and IITB-Monash Research Academy for financial support.

\clearpage

\onecolumn

\begin{center}
    \LARGE\bfseries Supplementary Material
\end{center}

\setcounter{section}{0}

\section{Ablation Study: Effect of Different VL backbones}

In addition to CLIP ViT-B, we have experimented with CLIP ViT-L as our VL backbone. Furthermore, we have explored two Remote Sensing-specific CLIP models: GeoRSCLIP (ViT-L) \cite{georsclip} and SkyCLIP (ViT-L) \cite{skyclip}. Table-\ref{abl_vlb} presents the quantitative results on the iSAID dataset for different VL backbones. Among them, CLIP ViT-L achieves the best performance. Notably, despite being trained on Remote Sensing data, GeoRSCLIP and SkyCLIP do not outperform CLIP ViT-L. This can likely be attributed to the lower image resolution used during their training.

\begin{table}[!h]
\centering
\resizebox{0.45\textwidth}{!}{
\begin{tblr}{
  cells = {c},
  hline{1,6} = {-}{0.08em},
  hline{2} = {-}{},
}
\textbf{VL backbone} & \textbf{Architecture} & \textbf{s-mIoU} & \textbf{u-mIoU} & \textbf{h-mIoU} \\
CLIP         & ViT-B        & \textcolor{blue}{75.48}  & 51.46  & 61.20  \\
CLIP         & ViT-L        & \textcolor{red}{79.85}  & \textcolor{red}{66.80} & \textcolor{red}{72.74}  \\
GeoRSCLIP    & ViT-L        & 75.47  & 50.32  & 60.38  \\
SkyCLIP      & ViT-L        & 66.75  & \textcolor{blue}{57.79}  & \textcolor{blue}{61.95}  
\end{tblr}
}
\caption{Quantitative comparison between different VL backbones on iSAID dataset.}
\label{abl_vlb}
\end{table}

\section{Visualization of Refined Correlation feature maps}

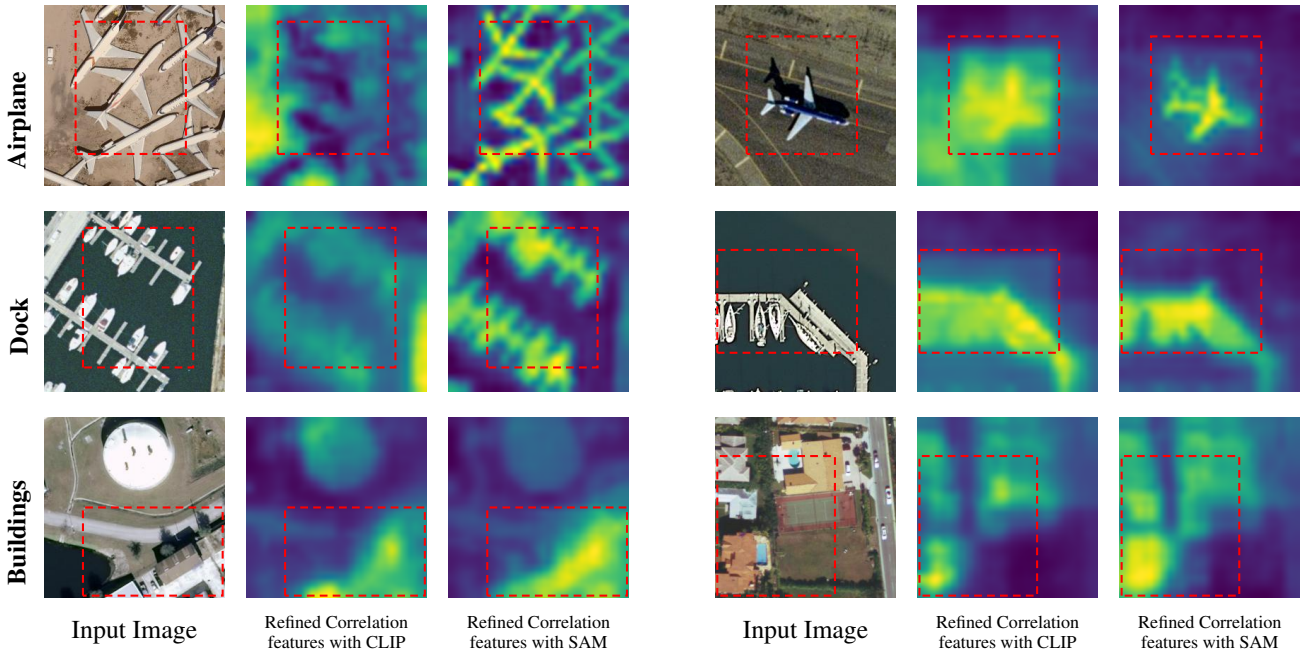
\begin{figure*}[!htbp]
    \centering
    \setlength{\tabcolsep}{1pt}
    \resizebox{1\textwidth}{!}{
    \begin{tabular}{@{}c@{\hskip 2pt} c@{\hskip 2pt} c@{\hskip 2pt} c@{\hskip 7pt} c@{\hskip 2pt} c@{\hskip 2pt} c@{}}
        \rotatebox{90}{\parbox{20mm}{\centering \textbf{Airplane}}} & 
        \begin{tikzpicture}
            \node at (0,0) {\includegraphics[width=0.14\textwidth]{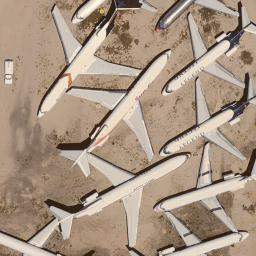}};
            \draw[red, thick, densely dashed] (-0.8,-0.8) rectangle (0.7,1);
        \end{tikzpicture} & 
        \begin{tikzpicture}
            \node at (0,0) {\includegraphics[width=0.14\textwidth]{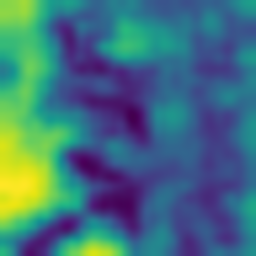}};
            \draw[red, thick, densely dashed] (-0.8,-0.8) rectangle (0.7,1);
        \end{tikzpicture} & 
        \begin{tikzpicture}
            \node at (0,0) {\includegraphics[width=0.14\textwidth]{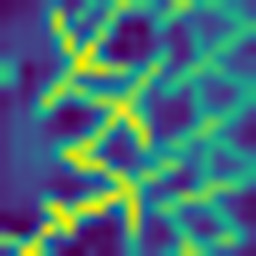}};
            \draw[red, thick, densely dashed] (-0.8,-0.8) rectangle (0.7,1);
        \end{tikzpicture} & 
        \qquad

        \begin{tikzpicture}
            \node at (0,0) {\includegraphics[width=0.14\textwidth]{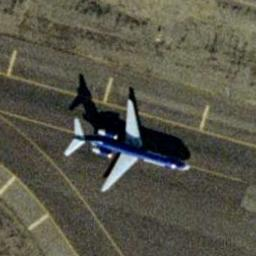}};
            \draw[red, thick, densely dashed] (-0.8,-0.8) rectangle (0.7,0.8);
        \end{tikzpicture} & 
        \begin{tikzpicture}
            \node at (0,0) {\includegraphics[width=0.14\textwidth]{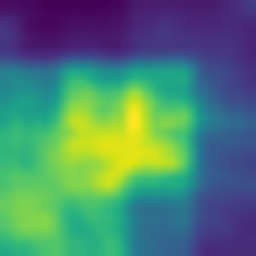}};
            \draw[red, thick, densely dashed] (-0.8,-0.8) rectangle (0.7,0.8);
        \end{tikzpicture} & 
        \begin{tikzpicture}
            \node at (0,0) {\includegraphics[width=0.14\textwidth]{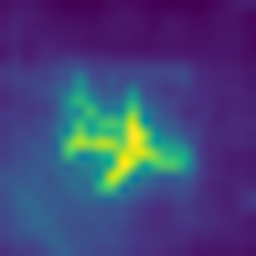}};
            \draw[red, thick, densely dashed] (-0.8,-0.8) rectangle (0.7,0.8);
        \end{tikzpicture} \\

        \rotatebox{90}{\parbox{20mm}{\centering \textbf{Dock}}} &

        \begin{tikzpicture}
            \node at (0,0) {\includegraphics[width=0.14\textwidth]{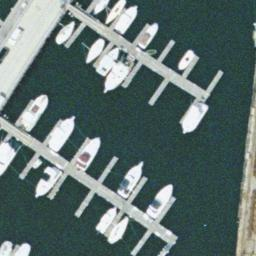}};
            \draw[red, thick, densely dashed] (-0.7,-0.9) rectangle (0.8,1);
        \end{tikzpicture} & 
        \begin{tikzpicture}
            \node at (0,0) {\includegraphics[width=0.14\textwidth]{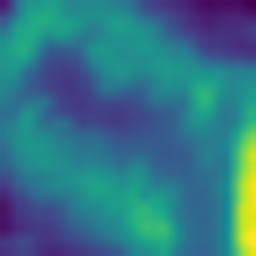}};
            \draw[red, thick, densely dashed] (-0.7,-0.9) rectangle (0.8,1);
        \end{tikzpicture} & 
        \begin{tikzpicture}
            \node at (0,0) {\includegraphics[width=0.14\textwidth]{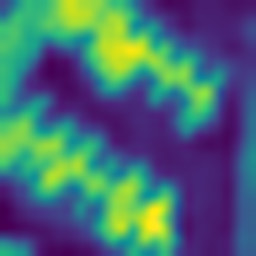}};
            \draw[red, thick, densely dashed] (-0.7,-0.9) rectangle (0.8,1);
        \end{tikzpicture} &
        \qquad
        \begin{tikzpicture}
            \node at (0,0) {\includegraphics[width=0.14\textwidth]{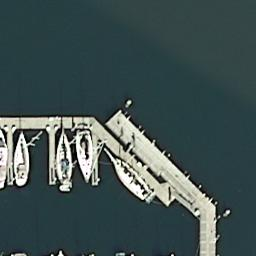}};
            \draw[red, thick, densely dashed] (-1.2,-0.7) rectangle (0.7,0.7);
        \end{tikzpicture} & 
        \begin{tikzpicture}
            \node at (0,0) {\includegraphics[width=0.14\textwidth]{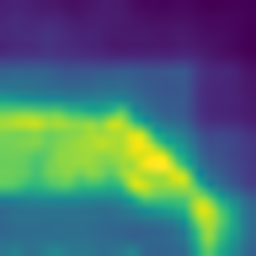}};
            \draw[red, thick, densely dashed] (-1.2,-0.7) rectangle (0.7,0.7);
        \end{tikzpicture} & 
        \begin{tikzpicture}
            \node at (0,0) {\includegraphics[width=0.14\textwidth]{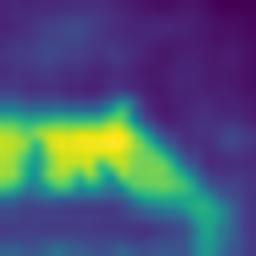}};
            \draw[red, thick, densely dashed] (-1.2,-0.7) rectangle (0.7,0.7);
        \end{tikzpicture} \\

        \rotatebox{90}{\parbox{20mm}{\centering \textbf{Buildings}}} & 
        \begin{tikzpicture}
            \node at (0,0) {\includegraphics[width=0.14\textwidth]{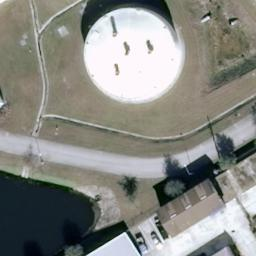}};
            \draw[red, thick, densely dashed] (-0.7,-1.2) rectangle (1.2,0);
        \end{tikzpicture} & 
        \begin{tikzpicture}
            \node at (0,0) {\includegraphics[width=0.14\textwidth]{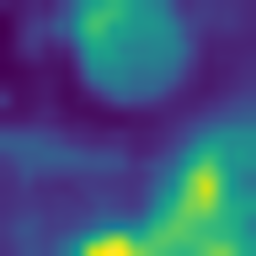}};
            \draw[red, thick, densely dashed] (-0.7,-1.2) rectangle (1.2,0);
        \end{tikzpicture} & 
        \begin{tikzpicture}
            \node at (0,0) {\includegraphics[width=0.14\textwidth]{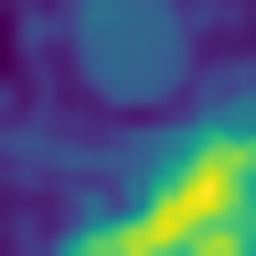}};
            \draw[red, thick, densely dashed] (-0.7,-1.2) rectangle (1.2,0);
        \end{tikzpicture} &
        \qquad
        \begin{tikzpicture}
            \node at (0,0) {\includegraphics[width=0.14\textwidth]{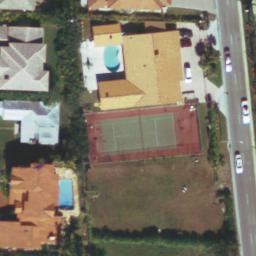}};
            \draw[red, thick, densely dashed] (-1.2,-1.2) rectangle (0.4,0.7);
        \end{tikzpicture} & 
        \begin{tikzpicture}
            \node at (0,0) {\includegraphics[width=0.14\textwidth]{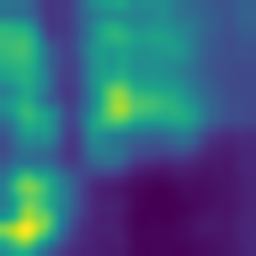}};
            \draw[red, thick, densely dashed] (-1.2,-1.2) rectangle (0.4,0.7);
        \end{tikzpicture} & 
        \begin{tikzpicture}
            \node at (0,0) {\includegraphics[width=0.14\textwidth]{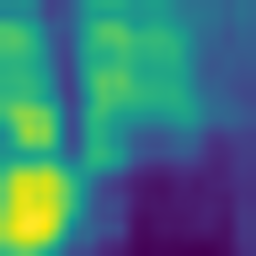}};
            \draw[red, thick, densely dashed] (-1.2,-1.2) rectangle (0.4,0.7);
        \end{tikzpicture} \\

        & \parbox{25mm}{\centering {Input Image}} & 
        \parbox{25mm}{\centering \scriptsize{Refined Correlation features with CLIP}} & 
        \parbox{25mm}{\centering \scriptsize{Refined Correlation features with SAM}} & 
        \qquad
        \parbox{25mm}{\centering {Input Image}} & 
        \parbox{25mm}{\centering \scriptsize{Refined Correlation features with CLIP}} & 
        \parbox{25mm}{\centering \scriptsize{Refined Correlation features with SAM}}  \\
    \end{tabular}
    }
    \caption{Refined Correlation feature visualization with CLIP and SAM guidance for different classes. SAM-refined correlation features can better discern between foreground and background for respective categories.}
    \label{fig:grid}
\end{figure*}

We visualize refined correlation maps for different classes, comparing the use of CLIP features versus SAM features as guidance in the Spatial Refinement block. Specifically, Fig. \ref{fig:grid} presents refined correlation features of different object categories for the Baseline and Config-A models (as discussed in Sec. 4.5 of main text). The results show that when SAM refines the correlation features, the refined features achieve better localization of the object of interest. 

\section{Additional comparison with state-of-the-art}

\subsection{Comparison with SegEarth-OV}

\begin{figure}[h]
    \centering
    \setlength{\tabcolsep}{1pt}
    \resizebox{0.75\textwidth}{!}{
    \begin{tabular}{c c c c}
        \begin{tikzpicture}
            \node at (0,0) {\includegraphics[width=0.2\textwidth]{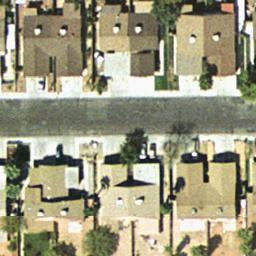}};
            \draw[red, thick, densely dashed] (-1.2,-1.5) rectangle (1.2,0.5);
        \end{tikzpicture} &
        \begin{tikzpicture}
            \node at (0,0) {\includegraphics[width=0.2\textwidth]{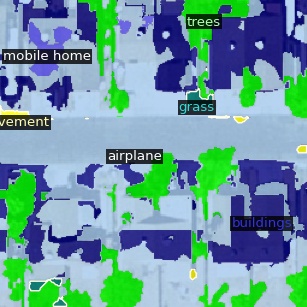}};
            \draw[red, thick, densely dashed] (-1.2,-1.5) rectangle (1.2,0.5);
        \end{tikzpicture} &
        \begin{tikzpicture}
            \node at (0,0) {\includegraphics[width=0.2\textwidth]{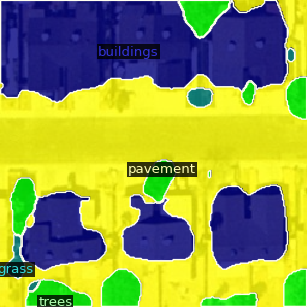}};
            \draw[red, thick, densely dashed] (-1.2,-1.5) rectangle (1.2,0.5);
        \end{tikzpicture} &
        \begin{tikzpicture}
            \node at (0,0) {\includegraphics[width=0.2\textwidth]{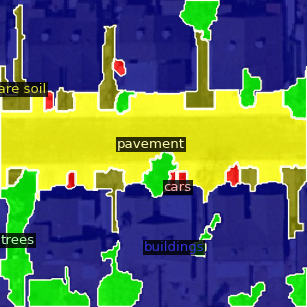}};
            \draw[red, thick, densely dashed] (-1.2,-1.5) rectangle (1.2,0.5);
        \end{tikzpicture} \\

        \begin{tikzpicture}
            \node at (0,0) {\includegraphics[width=0.2\textwidth]{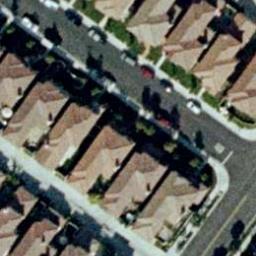}};
            \draw[red, thick, densely dashed] (-0.5,-0.5) rectangle (1.2,1.);
        \end{tikzpicture} &
        \begin{tikzpicture}
            \node at (0,0) {\includegraphics[width=0.2\textwidth]{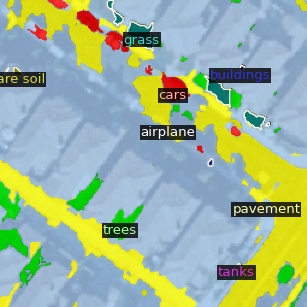}};
            \draw[red, thick, densely dashed] (-0.7,-0.7) rectangle (1,1);
        \end{tikzpicture} &
        \begin{tikzpicture}
            \node at (0,0) {\includegraphics[width=0.2\textwidth]{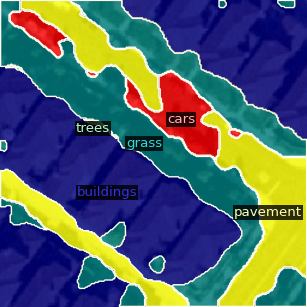}};
            \draw[red, thick, densely dashed] (-0.7,-0.7) rectangle (1,1);
        \end{tikzpicture} &
        \begin{tikzpicture}
            \node at (0,0) {\includegraphics[width=0.2\textwidth]{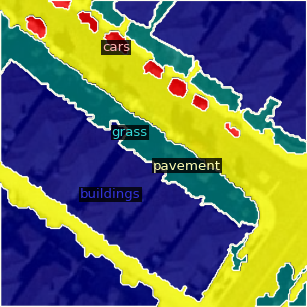}};
            \draw[red, thick, densely dashed] (-0.7,-0.7) rectangle (1,1);
        \end{tikzpicture} \\

        \subcaptionbox{Input Image}{
            \begin{tikzpicture}
                \node at (0,0) {\includegraphics[width=0.2\textwidth]{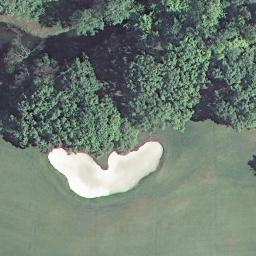}};
                \draw[red, thick, densely dashed] (-1.3,-1.3) rectangle (0.8,0.4);
            \end{tikzpicture}
        } &
        \subcaptionbox{SegEarth-OV}{
            \begin{tikzpicture}
                \node at (0,0) {\includegraphics[width=0.2\textwidth]{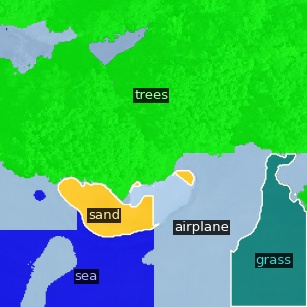}};
                \draw[red, thick, densely dashed] (-1.3,-1.3) rectangle (0.8,0.4);
            \end{tikzpicture}
        } &
        \subcaptionbox{\ourmodel\ (Ours)}{
            \begin{tikzpicture}
                \node at (0,0) {\includegraphics[width=0.2\textwidth]{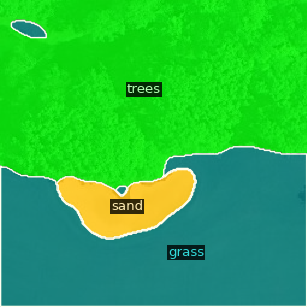}};
                \draw[red, thick, densely dashed] (-1.3,-1.3) rectangle (0.8,0.4);
            \end{tikzpicture}
        } &
        \subcaptionbox{Ground Truth}{
            \begin{tikzpicture}
                \node at (0,0) {\includegraphics[width=0.2\textwidth]{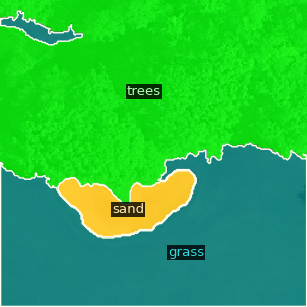}};
                \draw[red, thick, densely dashed] (-1.3,-1.3) rectangle (0.8,0.4);
            \end{tikzpicture}
        } \\
    \end{tabular}
    }
    \caption{Qualitative comparison between SegEarth-OV and our method. Dashed bounding boxes highlight regions where our model achieves more precise segmentation.}
    \label{segearth}
\end{figure}
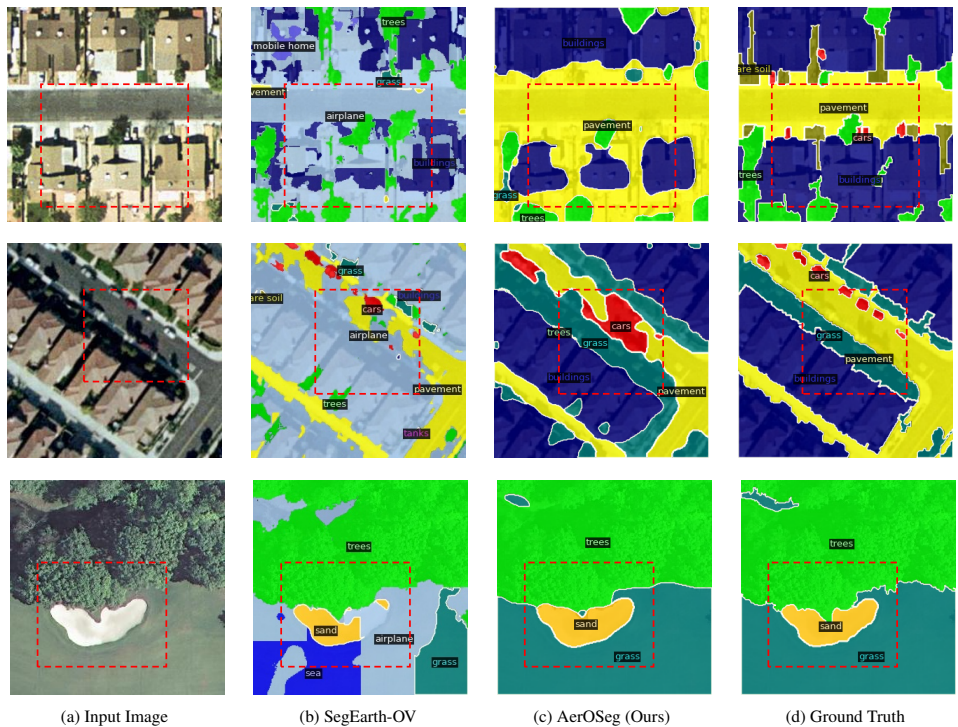

We have compared our model, AerOSeg with  state-of-the-art training-free method SegEarth-OV \cite{li2024segearth}. Since for training-free OVS, seen-unseen class split is irrelevant, we report mIoU scores over all classes. From Table~\ref{trainingfree}, we can see that our model performs significantly better than SegEarth-OV on all three datasets, highlighting the importance of domain-specific training. Fig. \ref{segearth} shows qualitative comparison between SegEarth-OV and our model. 

\begin{table}[!h]
\centering
\renewcommand{\arraystretch}{1.25}
\resizebox{0.4\linewidth}{!}{
\begin{tabular}{c|ccc}
\hline
\textbf{Methods}&\textbf{iSAID}&\textbf{DLRSD}&\textbf{OEM}\\
\hline
SegEarth-OV & 17.87& 18.92   & 29.72  \\
\textbf{AerOSeg (Ours) }&\textbf{65.87}&\textbf{51.62}& \textbf{49.71} \\
\hline
\end{tabular}
}
\caption{Quantitative comparison between training-free method SegEarth-OV and our proposed model.}
\label{trainingfree}
\end{table}

\subsection{More qualitative results}

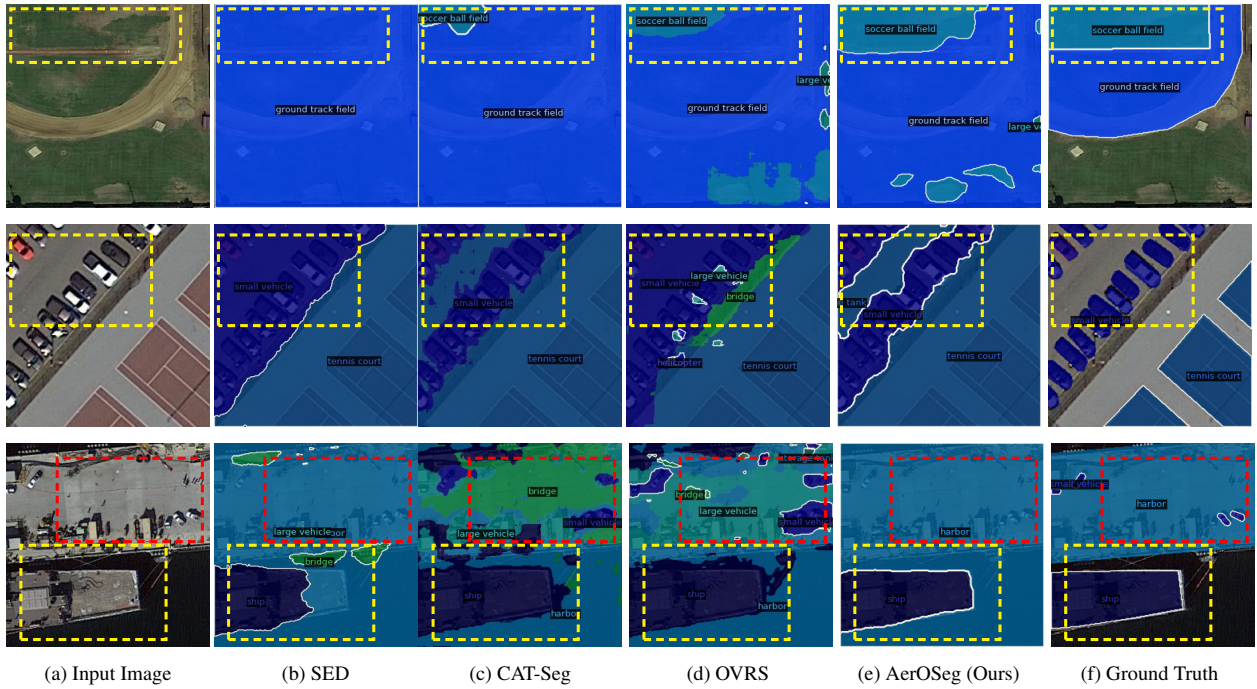
\begin{figure*}[!htbp]
    \centering
    \renewcommand{\arraystretch}{-5.0} 
    \setlength{\tabcolsep}{-3pt}
    \resizebox{0.97\textwidth}{!}{
    \begin{tabular}{@{}cccccc@{}}

        \begin{tikzpicture}
        \node at (0,0){\includegraphics[width=0.16\textwidth]{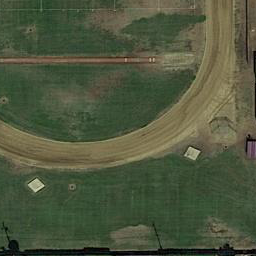}}; 
        \draw[yellow, very thick, densely dashed] (-1.34,0.6) rectangle (1,1.34);
        \end{tikzpicture} &
         \begin{tikzpicture}
        \node at (0,0){\includegraphics[width=0.16\textwidth]{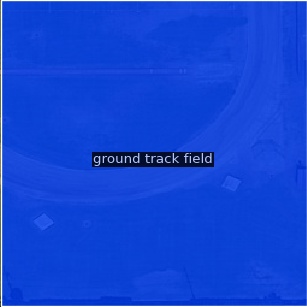}}; 
        \draw[yellow,very thick, densely dashed] (-1.34,0.6) rectangle (1,1.34);
        \end{tikzpicture} &
         \begin{tikzpicture}
        \node at (0,0){\includegraphics[width=0.16\textwidth]{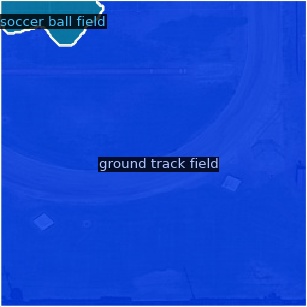}}; 
        \draw[yellow,very thick, densely dashed] (-1.34,0.6) rectangle (1,1.34);
        \end{tikzpicture} &
         \begin{tikzpicture}
        \node at (0,0){\includegraphics[width=0.16\textwidth]{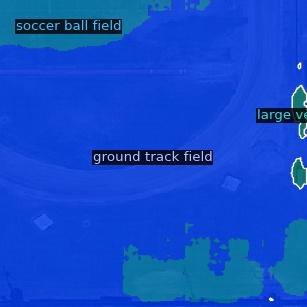}}; 
        \draw[yellow,very thick, densely dashed] (-1.34,0.6) rectangle (1,1.34);
        \end{tikzpicture} &
         \begin{tikzpicture}
        \node at (0,0){\includegraphics[width=0.16\textwidth]{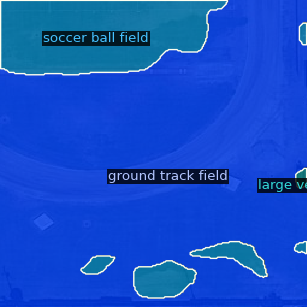}}; 
        \draw[yellow, very thick, densely dashed] (-1.34,0.6) rectangle (1,1.34);
        \end{tikzpicture} &
         \begin{tikzpicture}
        \node at (0,0){\includegraphics[width=0.16\textwidth]{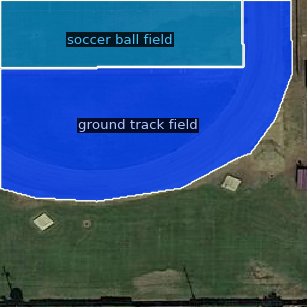}}; 
        \draw[yellow,very thick, densely dashed] (-1.34,0.6) rectangle (1,1.34);
        \end{tikzpicture} 
         
        \\

        \begin{tikzpicture}
        \node at (0,0){\includegraphics[width=0.16\textwidth]{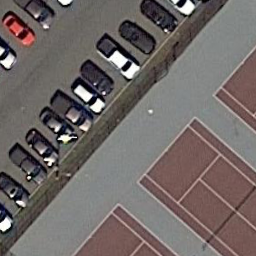}}; 
        \draw[yellow,very thick, densely dashed] (-1.34,0) rectangle (0.6,1.25);
        \end{tikzpicture} &
         \begin{tikzpicture}
        \node at (0,0){\includegraphics[width=0.16\textwidth]{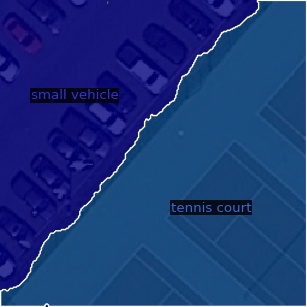}}; 
        \draw[yellow, very thick, densely dashed] (-1.34,0) rectangle (0.6,1.25);
        \end{tikzpicture} &
         \begin{tikzpicture}
        \node at (0,0){\includegraphics[width=0.16\textwidth]{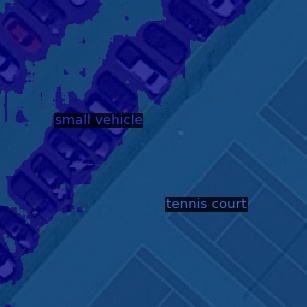}}; 
        \draw[yellow,very thick, densely dashed] (-1.34,0) rectangle (0.6,1.25);
        \end{tikzpicture} &
         \begin{tikzpicture}
        \node at (0,0){\includegraphics[width=0.16\textwidth]{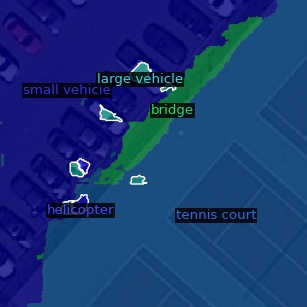}}; 
        \draw[yellow,very thick, densely dashed] (-1.34,0) rectangle (0.6,1.25);
        \end{tikzpicture} &
         \begin{tikzpicture}
        \node at (0,0){\includegraphics[width=0.16\textwidth]{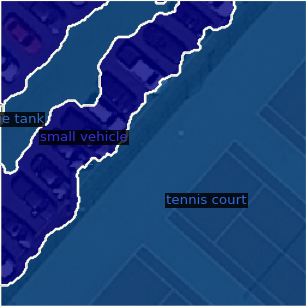}}; 
        \draw[yellow,very thick, densely dashed] (-1.34,0) rectangle (0.6,1.25);
        \end{tikzpicture} &
        \begin{tikzpicture}
        \node at (0,0){ \includegraphics[width=0.16\textwidth]{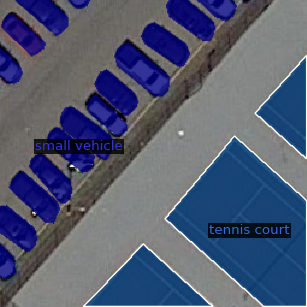}}; 
        \draw[yellow,very thick, densely dashed] (-1.34,0) rectangle (0.6,1.25);
        \end{tikzpicture}
        
        \\

        \subcaptionbox{Input Image}{\begin{tikzpicture}
        \node at (0,0){\includegraphics[width=0.16\textwidth]{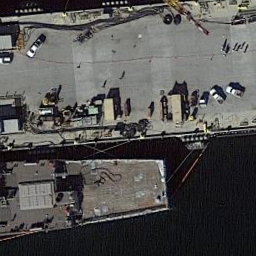}}; 
        \draw[yellow, very thick, densely dashed] (-1.2,-1.3) rectangle (0.8,0);
        \draw[red, very thick, densely dashed] (-0.7,0.05) rectangle (1.3,1.2);
        \end{tikzpicture} 
        }
        &
        \subcaptionbox{SED}{\begin{tikzpicture}
        \node at (0,0){\includegraphics[width=0.16\textwidth]{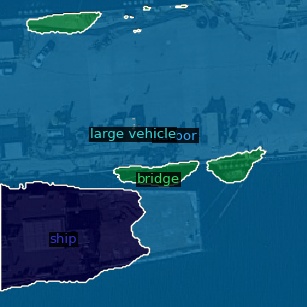}}; 
        \draw[yellow,very thick, densely dashed] (-1.2,-1.3) rectangle (0.8,0);
        \draw[red, very thick, densely dashed] (-0.7,0.05) rectangle (1.3,1.2);
        \end{tikzpicture}} &
        \subcaptionbox{CAT-Seg}
        {\begin{tikzpicture}
        \node at (0,0){\includegraphics[width=0.16\textwidth]{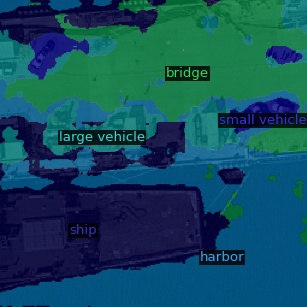}}; 
        \draw[yellow,very thick, densely dashed] (-1.2,-1.3) rectangle (0.8,0);
        \draw[red, very thick, densely dashed] (-0.7,0.05) rectangle (1.3,1.2);
        \end{tikzpicture}} &
        
        \subcaptionbox{OVRS}{
        \begin{tikzpicture}
        \node at (0,0){\includegraphics[width=0.16\textwidth]{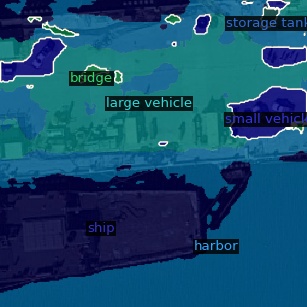}}; 
        \draw[yellow,very thick, densely dashed] (-1.2,-1.3) rectangle (0.8,0);
        \draw[red, very thick, densely dashed] (-0.7,0.05) rectangle (1.3,1.2);
        \end{tikzpicture}} &

        \subcaptionbox{\ourmodel\ (Ours)}{
       \begin{tikzpicture}
        \node at (0,0){ \includegraphics[width=0.16\textwidth]{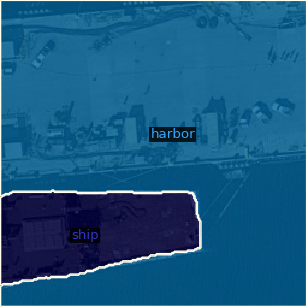}}; 
        \draw[yellow,very thick, densely dashed] (-1.2,-1.3) rectangle (0.8,0);
        \draw[red, very thick, densely dashed] (-0.7,0.05) rectangle (1.3,1.2);
        \end{tikzpicture}} &

        \subcaptionbox{Ground Truth}{
        \begin{tikzpicture}
        \node at (0,0){\includegraphics[width=0.16\textwidth]{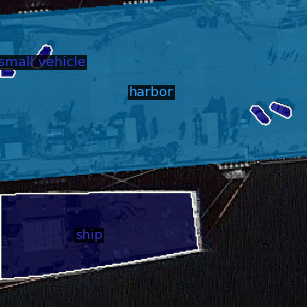}}; 
        \draw[yellow,very thick, densely dashed] (-1.2,-1.3) rectangle (0.8,0);
        \draw[red, very thick, densely dashed] (-0.7,0.05) rectangle (1.3,1.2);
        \end{tikzpicture}} 
         
    \end{tabular}
    }
    \caption{Qualitative comparison with state-of-the-art methods on iSAID dataset. Dashed bounding boxes highlight regions where our model achieves more precise segmentation.}
    \vspace{-15px}
    \label{isaid}
\end{figure*}

\begin{figure*}[!htbp]
    \centering
    \renewcommand{\arraystretch}{-5.0} 
    \setlength{\tabcolsep}{-3pt}
    \resizebox{0.97\textwidth}{!}{
    \begin{tabular}{@{}cccccc@{}}
        \begin{tikzpicture}
        \node at (0,0){\includegraphics[width=0.16\textwidth]{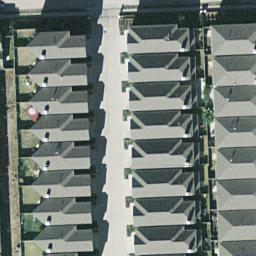}}; 
        \draw[green, very thick, densely dashed] (-0.05,-1.35) rectangle (0.9,1);
        \end{tikzpicture} &
        \begin{tikzpicture}
        \node at (0,0){\includegraphics[width=0.16\textwidth]{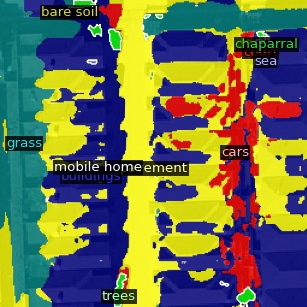}}; 
        \draw[green, very thick, densely dashed] (-0.05,-1.35) rectangle (0.9,1);
        \end{tikzpicture} &
        \begin{tikzpicture}
        \node at (0,0){\includegraphics[width=0.16\textwidth]{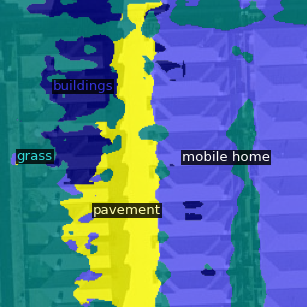}}; 
        \draw[green, very thick, densely dashed] (-0.05,-1.35) rectangle (0.9,1);
        \end{tikzpicture} &
        \begin{tikzpicture}
        \node at (0,0){\includegraphics[width=0.16\textwidth]{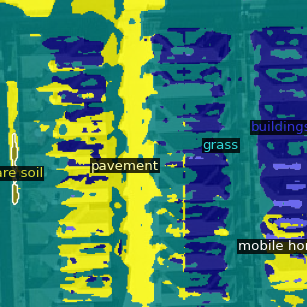}}; 
        \draw[green, very thick, densely dashed] (-0.05,-1.35) rectangle (0.9,1);
        \end{tikzpicture} &
        \begin{tikzpicture}
        \node at (0,0){\includegraphics[width=0.16\textwidth]{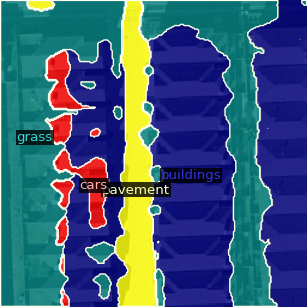}}; 
        \draw[green, very thick, densely dashed] (-0.05,-1.35) rectangle (0.9,1);
        \end{tikzpicture} &
        \begin{tikzpicture}
        \node at (0,0){\includegraphics[width=0.16\textwidth]{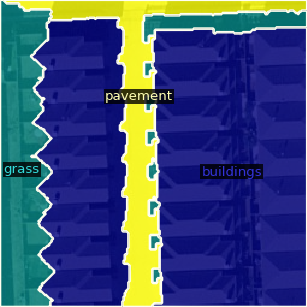}}; 
        \draw[green, very thick, densely dashed] (-0.05,-1.35) rectangle (0.9,1);
        \end{tikzpicture} \\

        \begin{tikzpicture}
        \node at (0,0){\includegraphics[width=0.16\textwidth]{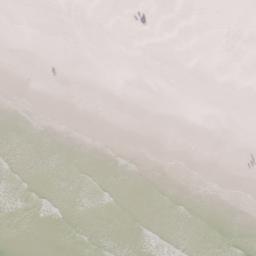}}; 
        \draw[red,very thick, densely dashed] (-0.7,-1.3) rectangle (1.3,0);
        \end{tikzpicture} &
        \begin{tikzpicture}
        \node at (0,0){\includegraphics[width=0.16\textwidth]{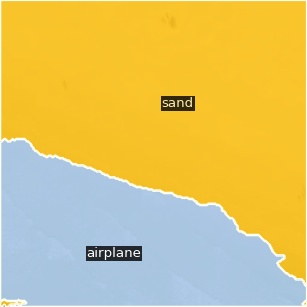}}; 
        \draw[red,very thick, densely dashed] (-0.7,-1.3) rectangle (1.3,0);
        \end{tikzpicture} &
        \begin{tikzpicture}
        \node at (0,0){\includegraphics[width=0.16\textwidth]{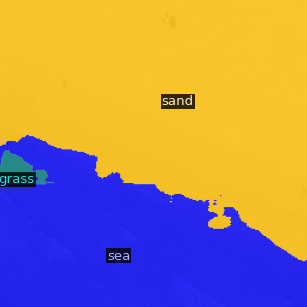}}; 
        \draw[red,very thick, densely dashed] (-0.7,-1.3) rectangle (1.3,0);
        \end{tikzpicture} &
        \begin{tikzpicture}
        \node at (0,0){\includegraphics[width=0.16\textwidth]{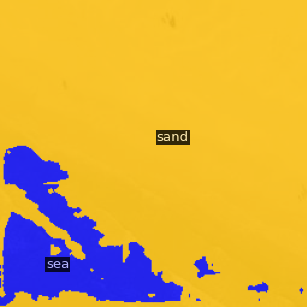}}; 
        \draw[red,very thick, densely dashed] (-0.7,-1.3) rectangle (1.3,0);
        \end{tikzpicture} &
        \begin{tikzpicture}
        \node at (0,0){\includegraphics[width=0.16\textwidth]{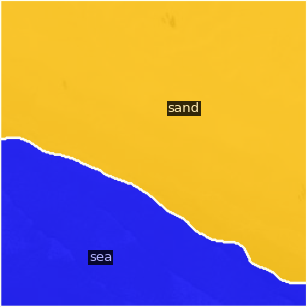}}; 
        \draw[red,very thick, densely dashed] (-0.7,-1.3) rectangle (1.3,0);
        \end{tikzpicture} &
        \begin{tikzpicture}
        \node at (0,0){\includegraphics[width=0.16\textwidth]{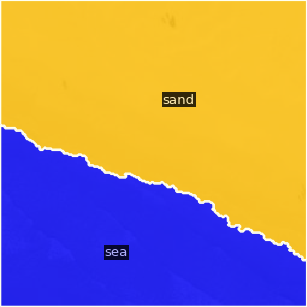}}; 
        \draw[red,very thick, densely dashed] (-0.7,-1.3) rectangle (1.3,0);
        \end{tikzpicture} 

        \\
        \subcaptionbox{Input Image}{ \begin{tikzpicture}
        \node at (0,0){
        \includegraphics[width=0.16\textwidth]{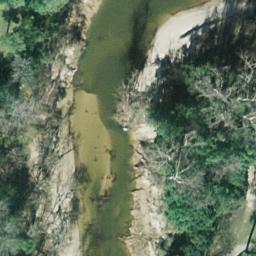}}; 
        \draw[red,very thick, densely dashed] (0.2,-1.3) rectangle (1.1,0);
        \draw[blue, very thick, densely dashed] (-1.34,-1.35) rectangle (-0.7,1.34);
        \end{tikzpicture}} &
        \subcaptionbox{SED}{
        \begin{tikzpicture}
        \node at (0,0){\includegraphics[width=0.16\textwidth]{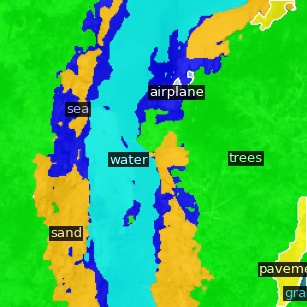}}; 
        \draw[red, very thick, densely dashed] (0.2,-1.3) rectangle (1.1,0);
        \draw[blue,  thick, densely dashed] (-1.34,-1.35) rectangle (-0.7,1.34);
        \end{tikzpicture}} &
        \subcaptionbox{CAT-Seg}{
        \begin{tikzpicture}
        \node at (0,0){\includegraphics[width=0.16\textwidth]{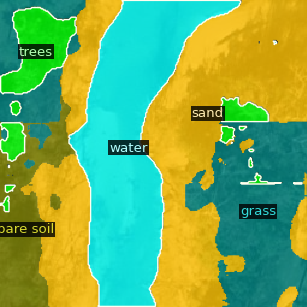}}; 
        \draw[red, very thick, densely dashed] (0.2,-1.3) rectangle (1.1,0);
        \draw[blue,thick, densely dashed] (-1.34,-1.35) rectangle (-0.7,1.34);
        \end{tikzpicture}} &
        \subcaptionbox{OVRS}{
        \begin{tikzpicture}
        \node at (0,0){\includegraphics[width=0.16\textwidth]{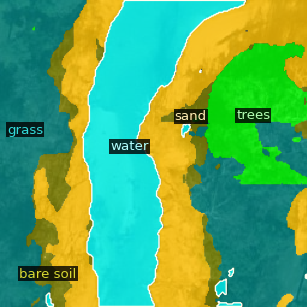}}; 
        \draw[red,very thick, densely dashed] (0.2,-1.3) rectangle (1.1,0);
        \draw[blue, thick, densely dashed] (-1.34,-1.35) rectangle (-0.7,1.34);
        \end{tikzpicture}} &
        \subcaptionbox{\ourmodel\ (Ours)}{
        \begin{tikzpicture}
        \node at (0,0){\includegraphics[width=0.16\textwidth]{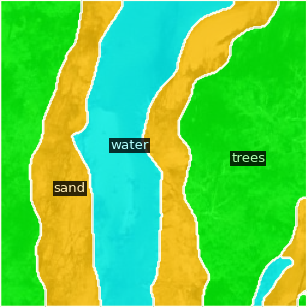}}; 
        \draw[red,very thick, densely dashed] (0.2,-1.3) rectangle (1.1,0);
        \draw[blue, thick, densely dashed] (-1.34,-1.35) rectangle (-0.7,1.34);
        \end{tikzpicture}} &
        \subcaptionbox{Ground Truth}{
        \begin{tikzpicture}
        \node at (0,0){\includegraphics[width=0.16\textwidth]{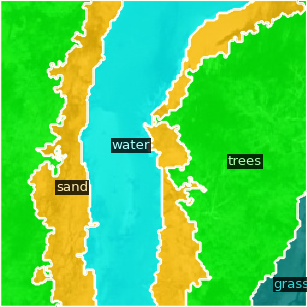}}; 
        \draw[red,very thick, densely dashed] (0.2,-1.3) rectangle (1.1,0);
        \draw[blue, thick, densely dashed] (-1.34,-1.35) rectangle (-0.7,1.34);
        \end{tikzpicture}} 
\end{tabular}
}
    \caption{Qualitative comparison with state-of-the-art methods on DLRSD dataset. Dashed bounding boxes highlight regions where our model achieves more precise segmentation.}
    \label{dlrsd}
    \vspace{-15px}
\end{figure*}

\begin{figure*}[t]
    \centering
    \renewcommand{\arraystretch}{-5.0} 
    \setlength{\tabcolsep}{-3pt}
    \resizebox{0.97\textwidth}{!}{
    \begin{tabular}{@{}cccccc@{}}
        \begin{tikzpicture}
        \node at (0,0){\includegraphics[width=0.16\textwidth]{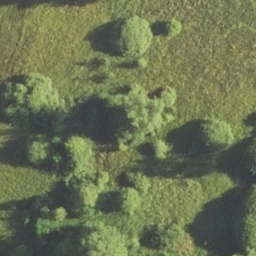}};
        \draw[red,very thick, densely dashed] (-0.45,-0.6) rectangle (1.2,0.6);
        \end{tikzpicture}&
        \begin{tikzpicture}
        \node at (0,0){\includegraphics[width=0.16\textwidth]{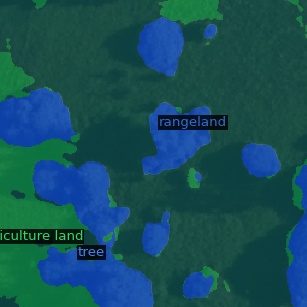}}; 
        \draw[red,very thick, densely dashed] (-0.45,-0.6) rectangle (1.2,0.6);
        \end{tikzpicture}&
        \begin{tikzpicture}
        \node at (0,0){\includegraphics[width=0.16\textwidth]{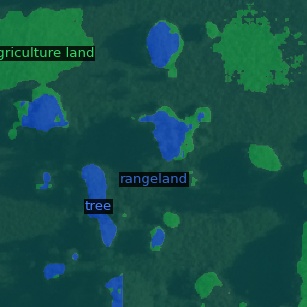}};
        \draw[red,very thick, densely dashed] (-0.45,-0.6) rectangle (1.2,0.6);
        \end{tikzpicture}&
        \begin{tikzpicture}
        \node at (0,0){\includegraphics[width=0.16\textwidth]{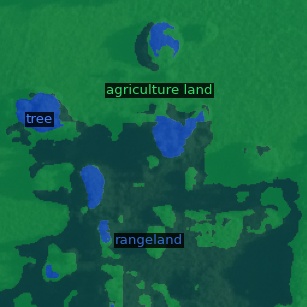}}; 
        \draw[red,very thick, densely dashed] (-0.45,-0.6) rectangle (1.2,0.6);
        \end{tikzpicture}&
        \begin{tikzpicture}
        \node at (0,0){ \includegraphics[width=0.16\textwidth]{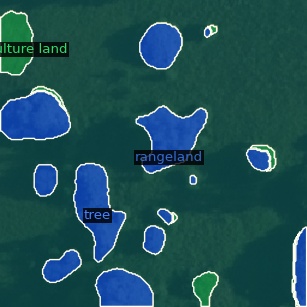}}; 
        \draw[red,very thick, densely dashed] (-0.45,-0.6) rectangle (1.2,0.6);
        \end{tikzpicture}&
        \begin{tikzpicture}
        \node at (0,0){\includegraphics[width=0.16\textwidth]{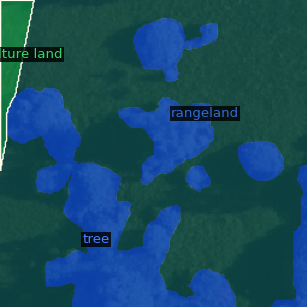}}; 
        \draw[red,very thick, densely dashed] (-0.45,-0.6) rectangle (1.2,0.6);
        \end{tikzpicture}\\
        
        \begin{tikzpicture}
        \node at (0,0){\includegraphics[width=0.16\textwidth]{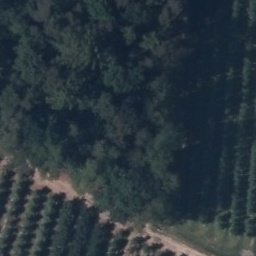}}; 
        \draw[red,very thick, densely dashed] (-0.7,-1.3) rectangle (1.3,0);
        \end{tikzpicture} &
        \begin{tikzpicture}
        \node at (0,0){\includegraphics[width=0.16\textwidth]{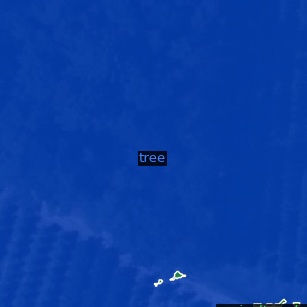}}; 
        \draw[red,very thick, densely dashed] (-0.7,-1.3) rectangle (1.3,0);
        \end{tikzpicture} &
        \begin{tikzpicture}
        \node at (0,0){\includegraphics[width=0.16\textwidth]{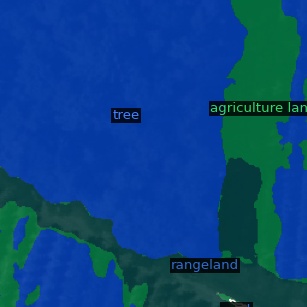}}; 
        \draw[red,very thick, densely dashed] (-0.7,-1.3) rectangle (1.3,0);
        \end{tikzpicture} &
        \begin{tikzpicture}
        \node at (0,0){\includegraphics[width=0.16\textwidth]{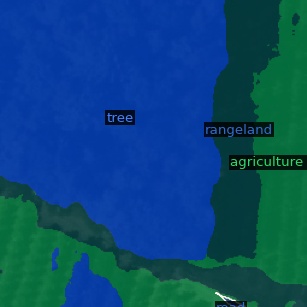}}; 
        \draw[red,very thick, densely dashed] (-0.7,-1.3) rectangle (1.3,0);
        \end{tikzpicture} &
        \begin{tikzpicture}
        \node at (0,0){\includegraphics[width=0.16\textwidth]{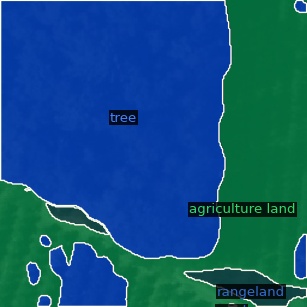}}; 
        \draw[red,very thick, densely dashed] (-0.7,-1.3) rectangle (1.3,0);
        \end{tikzpicture} &
        \begin{tikzpicture}
        \node at (0,0){\includegraphics[width=0.16\textwidth]{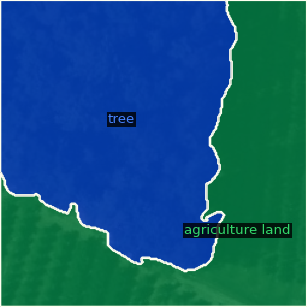}}; 
        \draw[red,very thick, densely dashed] (-0.7,-1.3) rectangle (1.3,0);
        \end{tikzpicture} \\

         \subcaptionbox{Input Image}{
        \begin{tikzpicture}
        \node at (0,0){\includegraphics[width=0.16\textwidth]{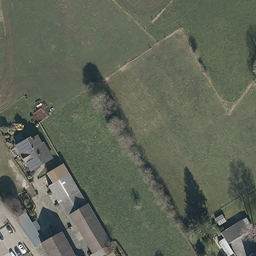}}; 
        \draw[red,very thick, densely dashed] (-0.6,-0.6) rectangle (0.6,1.0);
        \end{tikzpicture}} &
        \subcaptionbox{SED}{
        \begin{tikzpicture}
        \node at (0,0){\includegraphics[width=0.16\textwidth]{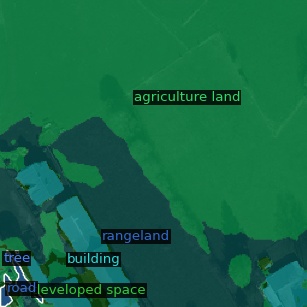}}; 
        \draw[red,very thick, densely dashed] (-0.6,-0.6) rectangle (0.6,1.0);
        \end{tikzpicture}} &
        \subcaptionbox{CAT-Seg}{
        \begin{tikzpicture}
        \node at (0,0){\includegraphics[width=0.16\textwidth]{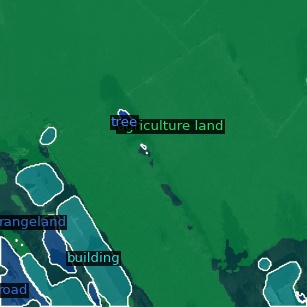}}; 
        \draw[red,very thick, densely dashed] (-0.6,-0.6) rectangle (0.6,1.0);
        \end{tikzpicture}} &
        \subcaptionbox{OVRS}{
        \begin{tikzpicture}
        \node at (0,0){\includegraphics[width=0.16\textwidth]{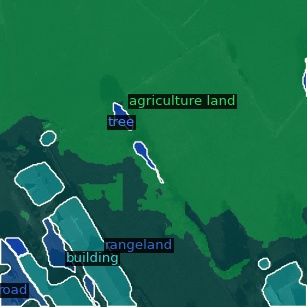}}; 
        \draw[red,very thick, densely dashed] (-0.6,-0.6) rectangle (0.6,1.0);
        \end{tikzpicture}} &
        \subcaptionbox{\ourmodel\ (Ours)}{
        \begin{tikzpicture}
        \node at (0,0){\includegraphics[width=0.16\textwidth]{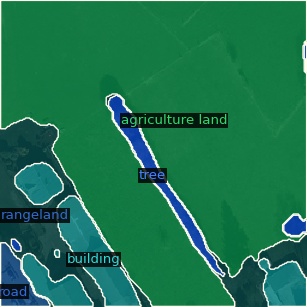}}; 
        \draw[red,very thick, densely dashed] (-0.6,-0.6) rectangle (0.6,1.0);
        \end{tikzpicture}} &
        \subcaptionbox{Ground Truth}{
        \begin{tikzpicture}
        \node at (0,0){\includegraphics[width=0.16\textwidth]{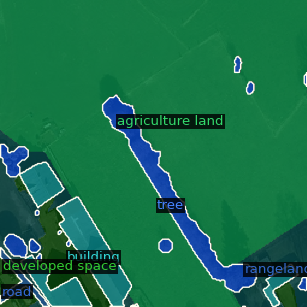}}; 
        \draw[red,very thick, densely dashed] (-0.6,-0.6) rectangle (0.6,1.0);
        \end{tikzpicture}} \\ 
     
\end{tabular}
}
    \caption{Qualitative comparison with state-of-the-art methods on OpenEarthMap dataset. Dashed bounding boxes highlight regions where our model achieves more precise segmentation.}
    \label{oem}
\end{figure*}

In this section, we have shown additional qualitative comparisons with various state-of-the-art models on iSAID, DLRSD and OpenEarthMap datasets in Fig. \ref{isaid}, \ref{dlrsd} and \ref{oem}, respectively.

\clearpage

{
    \small
    \bibliographystyle{ieeenat_fullname}
    \twocolumn 
    \bibliography{main}
}



\end{document}